\documentclass{article}

\usepackage[preprint]{neurips_2025}

\usepackage{amsmath,amssymb,amsfonts}
\usepackage{graphicx}
\usepackage{booktabs}
\usepackage{multirow}
\usepackage{xcolor}
\usepackage{hyperref}
\usepackage{url}
\usepackage{microtype}
\usepackage{caption}
\usepackage{subcaption}
\usepackage{enumitem}
\usepackage{algorithm}
\usepackage{algpseudocode}
\usepackage{tikz}

\hypersetup{
  colorlinks=true, linkcolor=blue, citecolor=blue, urlcolor=blue,
  pdftitle={CART: Context-Anchored Recurrent Transformer --- A Parameter-Efficient Architecture with Learned Stability},
  pdfauthor={Chad A. Capps},
  pdfsubject={Machine Learning; depth-recurrent transformer architecture for language modeling},
  pdfkeywords={depth-recurrent transformer, looped transformer, parameter efficiency, weight sharing, language modeling, ablation study, spectral radius, LTI stability}
}

\newcommand{\cart}{\textsc{CART}}

\newcommand{\pplwiki}{PPL$_\text{wiki}$}
\newcommand{\ppltiny}{PPL$_\text{tiny}$}
\newcommand{\ppledu}{PPL$_\text{edu}$}

\title{CART: Context-Anchored Recurrent Transformer\thanks{Code, training scripts, and the experiment database are available at \url{https://github.com/ccapps42/CART}.}\\
A Parameter-Efficient Architecture with Learned Stability}

\author{%
  Chad A. Capps\\
  Independent Researcher\\
  \texttt{Chad.Capps@me.com}
}

\begin{document}
\maketitle

\begin{abstract}
We present \cart{} (Context-Anchored Recurrent Transformer), a parameter-efficient language model architecture that reuses a single shared core block $R$ times across depth. Unlike prior looped transformers that recompute key-value representations at every loop iteration, \cart{} computes $K$ and $V$ once from a multi-layer prelude and reuses them throughout the recurrent core via multi-head latent attention (MLA) cross-attention. This design separates context encoding from iterative refinement, reduces per-loop FLOPs, and provides a stable attention anchor for all loop iterations. Stability of the recurrent computation is ensured by a learned Linear Time-Invariant (LTI) gate, which empirically settles on a narrow range of spectral radius values across all model scales ($\varrho \in [0.79, 0.83]$ at full training across all 36 configurations).

We evaluate \cart{} in two stages on single consumer GPUs (an RTX 3050 for Stage 1, an RTX 3090 for Stage 2). Stage 1 is a 64-configuration hyperparameter screen at 3{,}000 steps ($\sim$49M tokens) across $d \in \{256, 512, 768, 1024\}$, $R \in \{2, 4, 6, 8\}$, $P \in \{2, 3, 4, 6\}$. Stage 2 trains 36 configurations ($P{=}6$, $R \in \{6, 8, 10\}$, three seeds each) for 30{,}500 steps ($\sim$1B tokens at sequence length 1024). Stage 1 identifies $P{=}6$ as universally best and provides per-scale $R$ rankings; Stage 2 tests both at full training.

Stage 1's per-scale best-$R$ rankings reverse at Stage 2: Stage 1 ranks $R{=}8$ best at $d \in \{512, 768, 1024\}$ and $R{=}6$ best at $d{=}256$, while Stage 2 finds $R{=}6$ best at $d \geq 512$ and $R{=}8$ best at $d{=}256$ -- flipping all four per-scale predictions. Variable-$R$ inference at the best $d{=}1024$ model further shows that running more iterations than trained consistently degrades performance on HellaSwag, LAMBADA, and PIQA, a clean negative result for test-time compute scaling under our recipe. We attribute these reversals to the LTI gate settling into an aggressive-discard regime ($\varrho^R \approx 0.16$ at $d{=}1024$, $R{=}10$) and to a text-only corpus that may lack iteration-rewarding content. At the binding $d{=}1024$ parameter-parity comparison, \cart{} loses to a parameter-matched Dense baseline by $1$--$2\%$ on stored-parameter parity and by $\sim 10\%$ on effective-parameter parity. Diagnostic ablations decompose the larger effective-parameter gap into $\sim 5\%$ from weight sharing and a residual $\sim 5\%$ from the heterogeneous prelude $\to$ anchor $\to$ core $\to$ coda framing itself; the recurrent-core machinery components are individually vestigial.
\end{abstract}

\section{Introduction}
\label{sec:intro}

The dominant paradigm in language model architecture is the stacked transformer~\citep{vaswani2017attention}: $L$ unique layers, each with independent parameters, applied sequentially. This design scales predictably with $L$, but stored parameter count grows linearly with depth. At inference, every unique layer must be loaded from memory, making model size the primary bottleneck for deployment on memory-constrained hardware.

An alternative approach, depth-recurrent or looped transformers, reuses a single set of parameters across multiple passes through a shared block. The Universal Transformer~\citep{dehghani2018universal} pioneered this direction: a single block applied $R$ times to a token sequence. If the $R$-th application produces a representation of comparable quality to a $R$-layer unique model, then the recurrent model has achieved $R$-fold parameter compression at the cost of sequential compute.

The core challenge is making weight-sharing work. Naive looped transformers underperform their depth-matched counterparts~\citep{zeitoun2026hyperloop}, because a single block applied identically $R$ times cannot adapt its behavior across loop iterations. Recent solutions include hyper-connections~\citep{zhang2024hyperconnections,zeitoun2026hyperloop,schwethelm2026isodepth}, loop-index embeddings~\citep{su2024roformer,gomez2026openmythos}, and explicit stability constraints~\citep{prairie2026parcae,gomez2026openmythos}. The ``prelude / shared-weight core / coda'' topology used in this paper is also shared by several concurrent works~\citep{geiping2025latent,prairie2026parcae,gomez2026openmythos}; what distinguishes \cart{} is how the prelude representation enters the loop. All prior work in this family injects the prelude embedding $e$ via addition or concatenation into the running state, which then \emph{self-attends}. \cart{} instead computes $K, V$ once from $e$ and has the loop body \emph{cross-attend} to those frozen tensors.

\paragraph{Architectural contributions.}
\cart{} incorporates the proven mechanisms above (hyper-connections, loop-index embeddings, MLA compression) and contributes three architectural elements that, in combination, are not present in any prior or concurrent work we have surveyed:

\begin{enumerate}[leftmargin=*, topsep=2pt, itemsep=1pt]
  \item \textbf{Once-computed KV anchor with shared-weight cross-attending core.} The prelude computes $K, V$ once via MLA compression and holds them fixed across all $R$ core iterations. All surveyed concurrent looped architectures (Geiping~\citep{geiping2025latent}, Hyperloop~\citep{zeitoun2026hyperloop}, Parcae~\citep{prairie2026parcae}, OpenMythos~\citep{gomez2026openmythos}, and SpiralFormer~\citep{spiralformer2026}) recompute $K, V$ from the evolving hidden state at each loop step. The closest precedent for shared-weight iterated cross-attention is the Perceiver family~\citep{jaegle2021perceiver,jaegle2022perceiverio}, which re-projects $K, V$ from the input array at each iteration rather than caching the projected tensors and is non-autoregressive. \cart{}'s frozen anchor separates context encoding from iterative refinement, reduces per-loop FLOPs by eliminating redundant KV projections, and provides a stable attention target throughout recurrence.
  \item \textbf{Sigmoid-gated learned stability.} A sigmoid-gated LTI filter guarantees spectral radius $\varrho < 1$ without structural constraints: $h \leftarrow \sigma(a) \odot h_\text{in} + f_\text{transformer}(h)$. Parcae~\citep{prairie2026parcae} and OpenMythos~\citep{gomez2026openmythos} impose stability architecturally via zero-order-hold (ZOH) discretization of a state matrix; \cart{}'s gate instead \emph{learns} its convergence rate from data.
  \item \textbf{Causal cross-attention requirement.} In autoregressive recurrent models where prelude and core share the same token sequence, non-causal cross-attention leaks future tokens to the recurrent state. We document this constraint and its empirical consequences (Section~\ref{sec:arch_notes}).
\end{enumerate}

\paragraph{Empirical findings.}
This paper reports four findings about looped transformers that, to our knowledge, have not been demonstrated in prior work:

\begin{enumerate}[leftmargin=*, topsep=2pt, itemsep=1pt]
  \item \textbf{Learned memory timescale at full training.} Across all 36 Stage 2 configurations ($d \in \{256, 512, 768, 1024\}$ with $R \in \{6, 8, 10\}$, $P{=}6$, three seeds each, $\sim$1B tokens), the LTI gate's spectral radius settles in a narrow band, $\varrho \in [0.79, 0.83]$. Within this band, $\varrho$ rises monotonically with both $d$ ($+0.033$ from $d{=}256$ to $d{=}1024$ at $R{=}6$) and $R$ ($+0.008$ to $+0.009$ per pair of additional loops at fixed $d$). Stage 1 (Section~\ref{sec:stage1}) reported a tighter ``universal $\varrho \approx 0.893$'' across 64 configurations at step 3{,}000; that value is a mid-training transient, and $\varrho$ continues to decay through full training (Section~\ref{sec:spectral}).
  \item \textbf{Prelude depth dominates loop count.} The hyperparameter ordering $P{=}6 > P{=}4 > P{=}3 > P{=}2$ holds without exception across all $R$ values and all four scales. The pattern, that representational depth in the encoder matters more than recurrence depth in the core, has not been quantified across scales in prior work.
  \item \textbf{The Stage 1 R-benefit-grows-with-scale trend reverses at Stage 2.} In Stage 1 (3{,}000 steps), the \pplwiki{} improvement from $R{=}2$ to $R{=}8$ at $P{=}6$ ranged from $-0.25\%$ at $d{=}256$ (slight regression) to $+5.24\%$ at $d{=}1024$, suggesting recurrence depth becomes more valuable as width grows. At Stage 2 (30{,}500 steps, $\sim 10\times$ the training budget), this trend reverses: $R{=}6$ is the best of $\{6, 8, 10\}$ at every scale $d \geq 512$, with $R{=}10$ regressing by up to $1\%$ relative to $R{=}6$ at $d{=}1024$. We discuss three candidate explanations in Section~\ref{sec:stage1} (prelude saturation, aggressive LTI-gate discard, and corpus composition lacking iteration-rewarding content); a planned extended-training experiment to 2.5B tokens is the cleanest test of whether $R$-benefit re-emerges under longer training.
  \item \textbf{Parameter-efficiency claim does not hold at $d{=}1024$.} At the binding parameter-parity test ($d{=}1024$, $\sim 1$B tokens), Dense 7L (stored-parameter-matched, 3-seed mean) beats \cart{} by $1$--$2\%$ on every perplexity metric; Dense 12L (effective-parameter-matched, 3-seed mean) beats \cart{} by $\sim 10\%$ on the natural-language sets. Diagnostic ablations apportion the gap as $\sim 1\%$ from recurrence beyond $R{=}1$, $\sim 5\%$ from shared weights, and a residual $\sim 5\%$ from the heterogeneous prelude $\to$ anchor $\to$ core $\to$ coda architectural framing; the recurrent-core machinery components (HyperConnection, LTI gate, LIE) are individually vestigial. The shared-weight leverage thesis predicted by the original design does not deliver parameter efficiency at this scale and training budget.
\end{enumerate}

This paper presents \cart{}'s design, the Stage 1 hyperparameter sweep across 64 configurations at four scales, the Stage 2 full-training sweep across 36 configurations (3 seeds each), the reversal between the two stages' $R$-rankings, the $d{=}1024$ parameter-matched Dense comparison that bounds \cart{}'s parameter efficiency, and a series of diagnostic ablations that decompose the $d{=}1024$ \cart{}--Dense gap into specific architectural mechanisms.

\section{Related Work}
\label{sec:related}

\paragraph{Universal and looped transformers.}
\citet{dehghani2018universal} introduced the Universal Transformer, applying a single block recurrently with adaptive computation time. Subsequent work explored stability, efficiency, and training dynamics. \citet{jeddi2026loopformer} introduced budget-conditioned looping with shortcut-consistency training. \citet{chen2026deeper} demonstrates that shared-weight blocks iterated in latent space exhibit sharp compositional generalization absent in standard transformers. \citet{han2026hierarchical} empirically validates that hierarchical recurrence (distinct boundary layers + looped core) outperforms flat uniform looping, which directly supports \cart{}'s prelude/coda design.

\paragraph{Geiping recurrent-depth latent reasoning.}
\citet{geiping2025latent} introduced the prelude/core/coda terminology adopted in this paper, scaling a recurrent-depth language model to 3.5B parameters with the layout $(\ell_p, \ell_r, \ell_c) = (2, 4, 2)$. Their core block is shared-weight and iterated, but the prelude embedding $e$ is reinjected at each iteration via concatenation through an adapter $A: \mathbb{R}^{2h} \to \mathbb{R}^h$ mapping $[s_{i-1}; e]$, after which standard causal self-attention recomputes $K, V$ from the resulting state. Stability is achieved via a sandwich-RMSNorm pattern rather than a learned gate. \cart{} adopts the same three-zone layout but replaces the concat-and-self-attend mechanism with a frozen-KV cross-attention anchor and a sigmoid LTI gate.

\paragraph{Perceiver-style architectures.}
\citet{jaegle2021perceiver,jaegle2022perceiverio} introduced shared-weight iterated cross-attention from a learned latent array to a fixed input array, the closest structural precedent for \cart{}'s shared-weight cross-attending core. Two differences are essential. First, Perceiver re-projects $K, V$ from the input at every cross-attention call (the projection matrices are weight-tied; the resulting tensors are not cached). \cart{} caches the projected $K, V$ tensors themselves. Second, Perceiver's ``input array'' is a separate modality (bytes, images) attended to by a learned latent set, whereas \cart{}'s prelude output is itself a transformer encoding of the same token sequence used by the recurrent core, and the architecture is fully autoregressive and causal.

\paragraph{Hyperloop Transformers.}
\citet{zeitoun2026hyperloop} propose a structurally similar three-zone architecture (begin/middle/end, 25/50/25\% parameter split) with hyper-connections at loop boundaries. The critical difference from \cart{}: Hyperloop's middle block uses standard self-attention, recomputing $K, V$ from the current hidden state at every loop iteration. \cart{} computes $K, V$ once from the prelude and reuses them, reducing per-loop FLOPs and separating context encoding from iterative refinement. Additionally, \cart{} provides an explicit spectral radius guarantee via the LTI gate, which Hyperloop does not.

\paragraph{Parcae.}
\citet{prairie2026parcae} independently arrive at the looped prelude/coda structure and address stability via negative-diagonal parameterization of a state transition matrix, discretized via zero-order hold. This architecturally constrains $\varrho < 1$. \cart{} learns stability via a sigmoid gate: $h \leftarrow \sigma(a) \odot h_\text{in} + f_\text{transformer}(h)$, where $\varrho$ emerges from training rather than being structurally imposed. Parcae uses variable recurrence depth (Poisson-sampled) and truncated BPTT; \cart{} uses fixed $R$ and full backpropagation, which \citet{schwethelm2026isodepth} show preserves higher effective loop value ($\phi = 0.65$ vs.\ $0.38$ for truncated).

\paragraph{OpenMythos.}
\citet{gomez2026openmythos} is an open-source reconstruction of a hypothesized recurrent-depth architecture combining MLA, hyper-connections, loop-index embeddings, and LTI-stable injection. It is the most architecturally similar prior work to \cart{}. Two differences are architecturally significant. First, OpenMythos recomputes $K, V$ from the evolving hidden state at each loop iteration; \cart{} computes $K, V$ once from the prelude and holds them fixed. Second, OpenMythos implements LTI stability via ZOH discretization of a learned state matrix (the same mechanism as Parcae), whereas \cart{}'s sigmoid gate learns its convergence rate without structural constraints. \cart{} was informed by OpenMythos and shares its component vocabulary, but diverges in both the KV treatment and the stability mechanism.

\paragraph{Cross-layer parameter sharing without recurrent semantics.}
ALBERT~\citep{lan2020albert} shares parameters across stacked transformer layers as a compression strategy. The architecture is otherwise a standard non-recurrent encoder: there is no fixed-point semantics, no LTI gate, no prelude/coda separation, and no notion of ``iterating'' the shared block. MobileLLM~\citep{liu2024mobilellm} extends this idea to sub-billion-parameter on-device language models, applying block-wise weight sharing in a stacked-decoder layout to improve quality per parameter at fixed parameter count, the same regime \cart{} targets, but again without recurrent semantics or a stability mechanism. \cart{} uses cross-layer weight sharing in the same mechanical sense as both works (a single block is reused $R$ times) but interprets the reuse as recurrence with an explicit context anchor and a stability guarantee.

\paragraph{State-space models and linear-attention recurrence.}
Mamba~\citep{gu2023mamba}, Mamba-2~\citep{dao2024mamba2}, and RWKV~\citep{peng2023rwkv} achieve infinite-context reasoning via recurrence over the \emph{sequence} dimension, with parallel-scan training algorithms that recover transformer-like throughput. \cart{}'s recurrence is over the \emph{depth} dimension: $R$ shared-weight passes over a fixed-length token sequence. The two are complementary rather than competing: SSMs trade attention's quadratic context cost for fixed-state memory, while \cart{} trades unique-layer parameter cost for sequential depth iteration. A natural future direction is combining the two (an SSM-based prelude feeding a \cart{}-style recurrent core), but this paper restricts attention to a transformer prelude with MLA self-attention.

\paragraph{Other concurrent variants.}
Three additional concurrent works share aspects of \cart{}'s design but diverge on the key mechanistic details. \citet{spiralformer2026} adopts the same three-zone topology with shared-weight loop and pre-/post-loop blocks, but uses multi-resolution recursion with self-attention on the evolving state. \citet{plt2025parallel} reuses $K, V$ across all loop iterations (the closest precedent for the once-computed-KV idea), but the $K, V$ are projected from the first iteration of the loop block itself, not from a unique prelude over the same tokens, and serve a sliding-window/global hybrid attention pattern rather than a recurrent state anchor. \citet{rdvit2026} applies LTI-stable state injection to vision (semantic segmentation) using a single shared block with optional MoE; the cross-modal nature and dense-prediction setting make a direct comparison difficult, but the LTI-stable framing parallels \cart{}'s. None of these works combine all of: a unique-layer prelude producing a frozen KV anchor, a shared-weight cross-attending core, and a learned sigmoid LTI gate.

\paragraph{Iso-depth scaling laws.}
\citet{schwethelm2026isodepth} measure the recurrence-equivalence exponent $\phi$, defined as how much one additional loop contributes relative to a unique layer. They find $\phi = 0.46$ for vanilla looped models, rising to $\phi = 0.65$ with hyper-connections. \cart{} incorporates hyper-connections; in Section~\ref{sec:discussion} we discuss what \cart{}'s results imply for $\phi$, presenting qualitative implications rather than a numeric fit, which would require Dense reference depths beyond the two we trained at $d{=}1024$.

\paragraph{MLA compression.}
\citet{liu2024deepseek} introduced Multi-head Latent Attention for KV compression in mixture-of-experts models. \cart{} applies MLA to the prelude's KV projection ($d_{kv} = d/4$), making the once-computed context anchor cheaper to produce and store.

\section{Architecture}
\label{sec:arch}

\begin{figure}[t]
  \centering
  \includegraphics[width=0.55\linewidth, height=0.55\textheight, keepaspectratio]{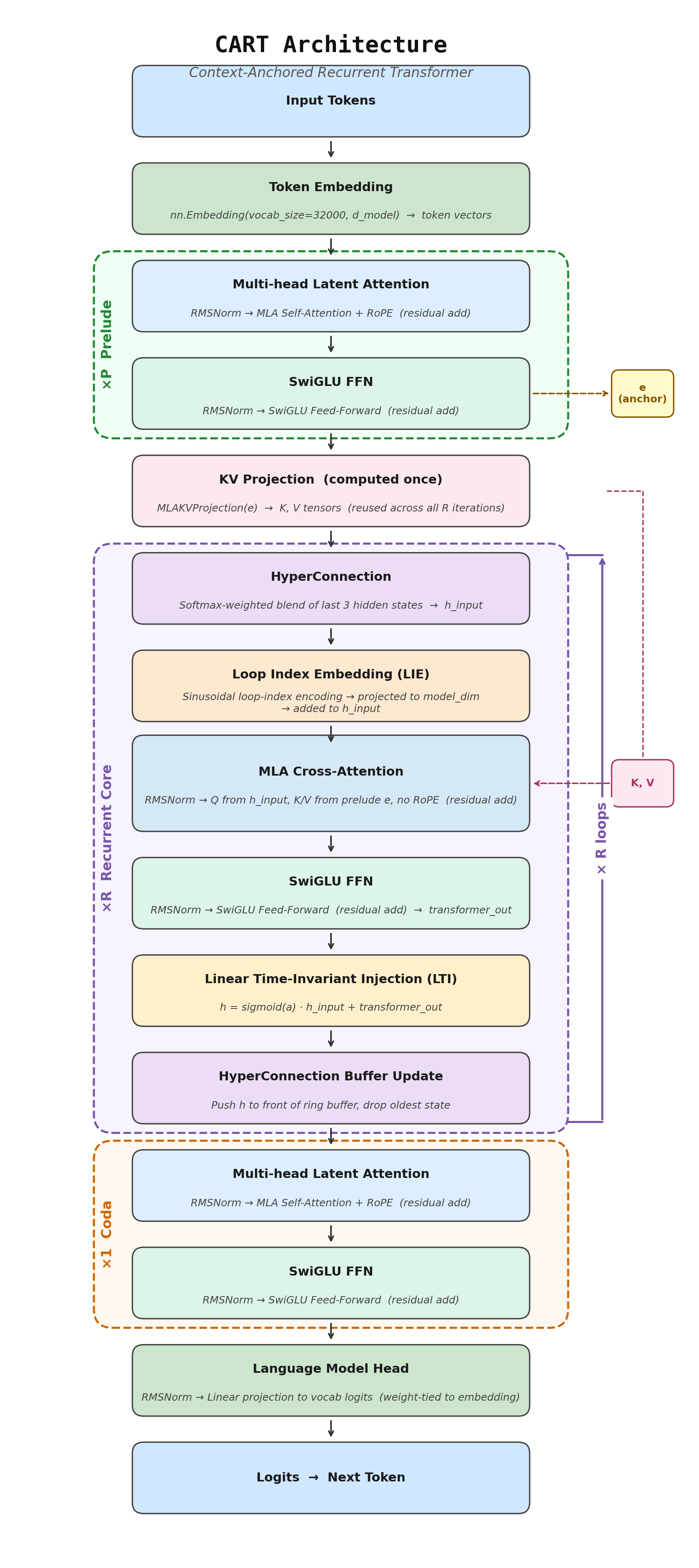}
  \caption{\cart{} architecture. The prelude produces context anchor $e$ via $P$ unique layers. KV projections are computed once from $e$ and reused across all $R$ core iterations. The coda produces final token representations.}
  \label{fig:arch}
\end{figure}

\cart{} divides depth into three sequential zones (Figure~\ref{fig:arch}):

\[
\text{Embedding} \to \underbrace{P \text{ unique layers}}_{\text{Prelude}} \to \underbrace{\text{KV}(e)}_{\text{once}} \to \underbrace{R \times \text{CoreBlock}}_{\text{Recurrent Core}} \to \underbrace{1 \text{ layer}}_{\text{Coda}} \to \text{Logits}
\]

\subsection{Prelude}

The prelude consists of $P$ unique transformer layers, each with MLA self-attention and SwiGLU FFN~\citep{shazeer2020glu}. MLA compresses the key-value cache: instead of projecting $K, V \in \mathbb{R}^{T \times d}$ directly, it first projects $x \in \mathbb{R}^{T \times d}$ to a compressed latent $c \in \mathbb{R}^{T \times d_{kv}}$ (with $d_{kv} = d/4$), then expands to full-rank $K, V$ for attention. The query $Q$ is full-rank. This reduces the cost of the KV projection by $4\times$.

The final prelude output $e \in \mathbb{R}^{T \times d}$ serves as the context anchor for the recurrent core. Before entering the loop, $K$ and $V$ are computed from $e$ once via a single MLA KV Projection module and held fixed for all $R$ iterations.

\subsection{Recurrent Core}

A single \texttt{CoreBlock} is applied $R$ times with shared weights. At each loop step $\ell \in \{1, \ldots, R\}$:

\begin{enumerate}[leftmargin=*, topsep=2pt, itemsep=1pt]
  \item \textbf{HyperConnection blend.} A ring buffer maintains the last $n_h = 3$ hidden states $\{h_{\ell-1}, h_{\ell-2}, h_{\ell-3}\}$. Softmax-weighted residual combination produces a blended input $\tilde{h}$, initialized with residual weights $[1, 0, 0]$~\citep{zhang2024hyperconnections}.

  \item \textbf{Loop Index Embedding (LIE).} A sinusoidal embedding of the loop index $\ell$ is added to $\tilde{h}$, providing a loop-step signal analogous to RoPE applied over recurrence depth rather than sequence position~\citep{su2024roformer,gomez2026openmythos}.

  \item \textbf{MLA cross-attention.} $Q$ is projected from $\tilde{h}$; $K, V$ are the prelude-derived cache from step 1. Attention is computed with \texttt{is\_causal=True} (see Section~\ref{sec:arch_notes}).

  \item \textbf{SwiGLU FFN.} Hidden dimension $d_{ff} = \lceil (8/3) \cdot d / 256 \rceil \times 256$ following standard practice.

  \item \textbf{LTI gate.} The output is passed through:
  \[
    h_\ell = \sigma(a) \odot \tilde{h} + f_\text{attn+FFN}(\tilde{h})
  \]
  where $a \in \mathbb{R}^d$ is a learned gate. The sigmoid gating ensures $0 < \sigma(a) < 1$, which, combined with the transformer output scaling, guarantees the spectral radius of the state transition satisfies $\varrho < 1$, ensuring convergence to a fixed point $h^*$ as $R \to \infty$~\citep{prairie2026parcae,gomez2026openmythos}.
\end{enumerate}

\subsection{Coda}

A single unique layer (MLA self-attention + SwiGLU FFN), structurally identical to a prelude layer. Its output feeds the language model head. The embedding matrix is tied to the output projection.

\subsection{Key Architectural Notes}
\label{sec:arch_notes}

\paragraph{Causal cross-attention requirement.}
In encoder-decoder transformers, non-causal cross-attention is correct because encoder and decoder process different sequences. In \cart{}, the prelude and recurrent core both operate on the same token sequence for autoregressive prediction. With non-causal cross-attention, $h[t]$ can attend to $e[t+1]$, which was built from token $t+1$, leaking the prediction target back into the state. We discovered this empirically: with $P=2$, no collapse occurred; with $P=3$, after approximately 1{,}750 training steps, the model discovered the leak and loss collapsed to near zero. The fix is \texttt{is\_causal=True} in \texttt{MLACrossAttention.forward()}. We report this as an architectural constraint for future work in this space.

\paragraph{Gradient checkpointing.}
Gradient checkpointing is disabled. The HyperConnection ring buffer creates tensor aliasing between loop iterations: $h_\text{input}$ is referenced both inside and outside the checkpoint boundary, and $K, V$ (identical objects) are passed across all $R$ iterations. Checkpointing with these aliasing patterns produces silent gradient corruption. VRAM for all tested configs ($d \leq 1024$, $R \leq 10$, seq\_len=1024) is well within 24 GB without checkpointing.

\paragraph{Fixed-point interpretation.}
The prelude output $e$ determines the fixed point $h^*$ that the recurrent core converges toward: $P$ sets the ceiling of representational quality, $R$ controls how closely $h^*$ is approached. This separation predicts that $P$ should dominate $R$, a prediction confirmed by the Stage 1 sweep.

\subsection{Parameter Count and Leverage}

For a config $(d, R, P)$, stored parameters are:
\[
N_\text{stored} = N_\text{prelude}(P) + N_\text{kv\_proj} + N_\text{core}(1) + N_\text{coda} + N_\text{embed}
\]
The core block appears in memory once but is \emph{applied} $R$ times, yielding effective parameters:
\[
N_\text{eff} = N_\text{stored} + (R-1) \cdot N_\text{core}
\]
Table~\ref{tab:params} shows stored and effective counts for selected configs.

\begin{table}[htbp]
\centering
\caption{Parameter counts for selected Stage 2 configs ($P=6$).}
\label{tab:params}
\footnotesize
\begin{tabular}{lrrr}
\toprule
Config & Stored & Effective & Leverage \\
\midrule
$d=256$, $R=6$ & 14.4M & 18.0M & 1.25$\times$ \\
$d=256$, $R=10$ & 14.4M & 21.4M & 1.49$\times$ \\
$d=512$, $R=6$ & 41.1M & 55.5M & 1.35$\times$ \\
$d=512$, $R=10$ & 41.1M & 76.5M & 1.86$\times$ \\
$d=768$, $R=6$ & 75.3M & 104.8M & 1.39$\times$ \\
$d=768$, $R=10$ & 75.3M & 146.3M & 1.94$\times$ \\
$d=1024$, $R=6$ & 125.1M & 178.8M & 1.43$\times$ \\
$d=1024$, $R=10$ & 125.1M & 221.8M & 1.77$\times$ \\
\bottomrule
\end{tabular}
\end{table}

\section{Experimental Setup}
\label{sec:setup}

\subsection{Training Data}

All models train on \texttt{stage2\_train.bin}, a single-pass mixed corpus of $\sim$1B tokens:
\begin{itemize}[leftmargin=*, topsep=2pt, itemsep=1pt]
  \item 30\% TinyStories~\citep{tinystories2023} (300M tokens, 1{,}024-token chunks)
  \item 30\% Wikipedia~\citep{wikidump} (300M tokens)
  \item 40\% FineWeb-Edu~\citep{fineweb2024} (400M tokens)
\end{itemize}

Interleaving uses 1{,}024-token chunks so every training window falls within a single source domain. The corpus is consumed in a single pass with no repetition. Tokenizer: Llama-2 (NousResearch/Llama-2-7b-hf, 32{,}000 vocab)~\citep{touvron2023llama}.

\subsection{Stage 1: Hyperparameter Screening}

Stage 1 trains each config for 3{,}000 steps ($\sim$49M tokens) at \texttt{seq\_len}=512, \texttt{batch}=4, \texttt{grad\_accum}=4 (16{,}384 tokens/step). Evaluation every 500 steps on three held-out sets:
\begin{itemize}[leftmargin=*, topsep=2pt, itemsep=1pt]
  \item \ppltiny{}: TinyStories validation (synthetic narrow-vocabulary text)
  \item \pplwiki{}: Wikipedia shard 40 (natural-language general text)
  \item \ppledu{}: FineWeb-Edu shard 97 (natural-language general text)
\end{itemize}
Stage 1 sweeps $d \in \{256, 512, 768, 1024\}$, $R \in \{2, 4, 6, 8\}$, $P \in \{2, 3, 4, 6\}$ (64 configs total), run on an RTX 3050 (8 GB).

\paragraph{A note on evaluation terminology.}
\label{sec:eval_terminology}
All three evaluation sets are held-out shards of the same corpora used in training; they are therefore strictly in-distribution under the standard machine-learning definition of distribution shift. We use the labels \emph{narrow-vocabulary} and \emph{natural-language} throughout the paper to distinguish a difference in baseline difficulty rather than a difference in distribution. TinyStories~\citep{tinystories2023} is synthetically generated text with a vocabulary of roughly 1{,}500 words and simple grammar: a small language model can fit it tightly, and absolute perplexities are correspondingly low (typically 3--7 in our sweep). Wikipedia and FineWeb-Edu are natural human-written text with unconstrained vocabulary and complex syntax; perplexity on these sets reflects general-domain language-modeling quality, with absolute values an order of magnitude higher (typically 25--250). When this paper reports that \cart{}'s advantage is ``larger on the natural-language sets,'' we mean that the gap to a parameter-matched Dense baseline grows when the evaluation target is harder, not that \cart{} generalizes to a held-out distribution. The standard out-of-distribution evaluation suite (cross-domain transfer, multilingual, code) is reserved for future work.

\subsection{Stage 2: Full Training}

Stage 2 trains selected configs (P=6, $R \in \{6, 8, 10\}$, 3 seeds each) for 30{,}500 steps ($\sim$1B tokens) at \texttt{seq\_len}=1024, \texttt{batch}=8, \texttt{grad\_accum}=4 (32{,}768 tokens/step). Optimizer: AdamW8bit~\citep{dettmers2022gptint8}, $\text{lr}_\text{max}=3 \times 10^{-4}$, cosine decay to $3 \times 10^{-5}$, 100-step warmup, weight decay 0.1, gradient clip 1.0. All runs on RTX 3090 (24 GB). No \texttt{torch.compile} (unavailable on Windows 11/no Triton).

Dense baselines (standard MHA + SwiGLU, $P{=}0$) are trained at $d{=}1024$ in two parameter-matched variants: a 7-layer Dense matching CART's \emph{stored} parameter count ($\sim 75$M) and a 12-layer Dense matching CART's \emph{effective} parameter count at $R{=}6$ ($\sim 105$M). Smaller-scale Dense baselines were not pursued in this study; $d{=}1024$ is the binding test of the parameter-efficiency claim. Both Dense variants use identical Stage 2 hyperparameters (3-seed sweep, same optimizer, same data, same $30{,}500$ steps).

\subsection{Reproducibility}
\label{sec:repro}

\paragraph{Random seeds.} All Stage 2 configurations are trained with three seeds: $\{42, 137, 271\}$. Reported perplexities are means across seeds. Per-seed standard deviations are tight at $d{=}512$ (max std on \pplwiki{}: 0.17 across all $R \in \{6, 8, 10\}$, or $0.6\%$ of the mean) and wider but still consistent at $d{=}256$ (max std: 3.06, or $7.5\%$ of the mean). The widening at small $d$ reflects loss-landscape sensitivity at low parameter counts rather than instability in the architecture; gradient norm remained bounded and no run diverged. The $d{=}1024$ Dense baselines exhibit tighter seed-to-seed agreement than CART at the same scale (7L Dense std on \pplwiki{} $= 0.06$ vs CART's $0.08$ at $R{=}6$; 12L Dense std $= 0.05$), supporting the interpretation that the CART$-$Dense gap reported in Tables~\ref{tab:stage2_results} and~\ref{tab:stage2_dense12l} reflects architectural signal rather than seed noise. Per-seed values are available in the released experiment database.

\paragraph{Hardware and compute budget.} Stage 1 (64 configurations) ran on a single RTX 3050 (8 GB), $\sim 1$--$3$ hours per configuration. Stage 2 (36 configurations) runs on a single RTX 3090 (24 GB), $\sim 2$--$13$ hours per configuration depending on $(d, R)$. The full Stage 2 sweep fits in $\sim 7.5$ days of single-GPU wall-clock time. No multi-node, no cloud compute. Total Stage 2 training compute for the full sweep was approximately $\mathbf{22}$ \textbf{ExaFLOPs} (using the standard $6N$ per-token estimate~\citep{kaplan2020scaling}, summed over all 36 runs at 1B tokens each), which is roughly $0.25$ PetaFLOP-days. Per-run compute ranges from $0.1$ ExaFLOP at $d{=}256, R{=}6$ to $1.3$ ExaFLOP at $d{=}1024, R{=}10$. For comparison, training a single GPT-3 small (125M parameters)~\citep{brown2020gpt3} on its 300B-token budget required approximately $2.25 \times 10^{20}$ FLOPs ($\sim 225$ ExaFLOPs, or $\sim 2.6$ petaflop/s-days under the same $6N$ estimate); the full \cart{} 36-config sweep across four scales is therefore roughly one-tenth the total compute of a single GPT-3-small training.

\paragraph{Training-data ordering.} The training corpus \texttt{stage2\_train.bin} is a fixed pre-shuffled file consumed in a single pass. Different seeds therefore see the same data in the same order; randomness affects only weight initialization and dropout-equivalent stochastic operations. This isolates seed sensitivity to model initialization rather than data order.

\paragraph{Token budget and Chinchilla compliance.} Stage 2 trains every configuration on $\sim 1$B tokens. \cart{}'s weight sharing makes Chinchilla compliance ambiguous: a single core block is stored once but applied $R$ times, so the stored parameter count and the effective parameter count differ by up to $1.8\times$ at our scales. We report both ratios because they answer different questions: \emph{stored} captures inference-time memory and HuggingFace download size; \emph{effective} captures compute-equivalent capacity and is the right denominator for predicting where training loss saturates (Chinchilla's $6N$ FLOPs-per-token derivation uses the effective compute count for a standard transformer).

\begin{table}[htbp]
\centering
\caption{Stage 2 Chinchilla compliance per $(d, R)$ at 1B tokens, measured against both stored and effective parameters. Chinchilla-optimal target is $20{:}1$.}
\label{tab:chinchilla}
\footnotesize
\begin{tabular}{rrrrrrr}
\toprule
$d$ & $R$ & Stored (M) & Effective (M) & Stored ratio & Effective ratio & \% Chinchilla (eff) \\
\midrule
 256 & 6  & 14.4  & 18.0  & 69:1 & 56:1 & 278\% (over) \\
 256 & 10 & 14.4  & 21.4  & 69:1 & 47:1 & 234\% (over) \\
 512 & 6  & 41.1  & 55.5  & 24:1 & 18:1 &  90\% (near) \\
 512 & 10 & 41.1  & 76.5  & 24:1 & 13:1 &  65\% (sub)  \\
 768 & 6  & 75.3  & 104.8 & 13:1 & 10:1 &  48\% (sub)  \\
 768 & 10 & 75.3  & 146.3 & 13:1 &  7:1 &  34\% (sub)  \\
1024 & 6  & 125.1 & 178.8 &  8:1 &  6:1 &  28\% (sub)  \\
1024 & 10 & 125.1 & 221.8 &  8:1 &  5:1 &  23\% (sub)  \\
\bottomrule
\end{tabular}
\end{table}

Two consequences follow. First, cross-scale comparisons in this paper are not measured at matched-Chinchilla budgets on either parameter count; $d \geq 768$ configurations are firmly sub-Chinchilla on both measures, and the $d{=}1024$ models in particular run at $23$--$28\%$ of optimal training. Their reported perplexities are lower bounds on quality, not asymptotic values. Second, claims about the limiting behavior of \cart{}'s recurrence (whether more $R$ keeps helping at full training, whether the spectral radius converges to a stable value at full training) cannot be made from Stage 2 data alone. The extended-training run (Section~\ref{sec:discussion}) is the test that addresses these questions.

For context, recent sub-1B language models trained for downstream deployment are typically far over-Chinchilla on both measures: TinyLlama-1.1B~\citep{zhang2024tinyllama} trains on 3T tokens ($\sim 2700{:}1$), Pythia-160M~\citep{biderman2023pythia} trains on 300B tokens ($\sim 1875{:}1$). Our $5{-}278\%$ range reflects the goals of an architectural sweep rather than a deployment-targeted release. We discuss this further in Limitations (Section~\ref{sec:discussion}).

\paragraph{Code, training scripts, and database.} The full implementation, all training and evaluation scripts, the SQLite database of every run (\texttt{results.db}), and the figure-generation scripts are released at the URL in the title footnote.

\subsection{FLOPs Budget}

Forward-pass FLOPs per sequence for representative Stage 2 configs at $d=1024$, $T=1024$ are shown in Table~\ref{tab:flops}. The per-optimizer-step cost is $3 \times N_\text{seq/step} \times F_\text{fwd}$ (backward $\approx 2\times$ forward).

\begin{table}[htbp]
\centering
\caption{Forward-pass GFLOPs per sequence at $d=1024$, $T=1024$.}
\label{tab:flops}
\footnotesize
\begin{tabular}{lrrr}
\toprule
Model & GFLOPs & Stored (M) & Eff.\ layers \\
\midrule
Dense 7L & 214 & 123 & 7.0 \\
CART $R{=}6$, $P{=}6$ & 355 & 125 & 11.6 \\
CART $R{=}8$, $P{=}6$ & 407 & 125 & 13.3 \\
CART $R{=}10$, $P{=}6$ & 460 & 125 & 15.0 \\
Dense 17L (FLOPs-eq.) & 520 & 251 & 17.0 \\
\bottomrule
\end{tabular}
\end{table}

CART $R=10$, $P=6$ delivers 15 Dense-layer-equivalent FLOPs from 125M stored parameters, compared to 251M stored for the FLOPs-matched 17-layer Dense (1.70$\times$ more storage for the same compute budget).

\subsection{Training Throughput and Hardware Utilization}
\label{sec:throughput}

Table~\ref{tab:throughput} reports measured Stage 2 training throughput on a single RTX 3090 (BF16 peak: 71 TFLOPS) across all $(d, R)$ combinations, averaged over three seeds per configuration. Throughput scales as expected with both model width and loop count: at $d{=}1024$ the throughput per token is roughly $1/4$ that at $d{=}256$, and each additional pair of loops costs $\sim 10$--$13\%$ of throughput at fixed $d$.

Model FLOPs Utilization (MFU)~\citep{chowdhery2023palm} measures the fraction of theoretical peak compute used during training. \cart{}'s MFU rises monotonically with model width, from $\sim 20\%$ at $d{=}256$ to $\sim 33$--$37\%$ at $d{=}512$, $\sim 40$--$45\%$ at $d{=}768$, and $\sim 48$--$49\%$ at $d{=}1024$. The improvement comes from larger matrix multiplications saturating Tensor Cores more effectively. For context, \citet{chowdhery2023palm} report MFU of $46.2\%$ for PaLM-540B on TPU v4, \citet{brown2020gpt3} estimates around $21\%$ for GPT-3 175B on V100s, and \citet{touvron2023llama} report $\sim 45\%$ for Llama-2 training on A100. \cart{}'s $\sim 49\%$ MFU at $d{=}1024$ on a single consumer RTX 3090 is within the upper range reported for transformer training and demonstrates that the depth-recurrent loop structure does not impose a hardware-utilization penalty at the larger scales. Parameter-matched Dense-baseline throughputs were not run in this study and are left for follow-up work.

\begin{table}[htbp]
\centering
\caption{Measured Stage 2 training throughput and Model FLOPs Utilization on a single RTX 3090. Effective batch is 32{,}768 tokens/step in all cases. MFU computed against BF16 peak of $71$ TFLOPS using the standard $6N$ FLOPs-per-token estimate where $N$ is the effective parameter count.}
\label{tab:throughput}
\footnotesize
\begin{tabular}{ccrrrr}
\toprule
$d$ & $R$ & Eff.\ params (M) & Tokens/sec & sec/step & MFU \\
\midrule
256  & 6  &  18.0 & 134{,}985 & 0.243 & 20\% \\
256  & 8  &  19.7 & 120{,}855 & 0.271 & 20\% \\
256  & 10 &  21.4 & 110{,}408 & 0.297 & 20\% \\
512  & 6  &  55.5 &  71{,}244 & 0.460 & 33\% \\
512  & 8  &  66.0 &  63{,}441 & 0.517 & 35\% \\
512  & 10 &  76.5 &  57{,}307 & 0.572 & 37\% \\
768  & 6  & 104.8 &  45{,}556 & 0.719 & 40\% \\
768  & 8  & 125.5 &  40{,}495 & 0.809 & 43\% \\
768  & 10 & 146.3 &  36{,}325 & 0.902 & 45\% \\
1024 & 6  & 178.8 &  32{,}393 & 1.012 & 49\% \\
1024 & 8  & 200.3 &  28{,}711 & 1.141 & 49\% \\
1024 & 10 & 221.8 &  25{,}681 & 1.276 & 48\% \\
\bottomrule
\end{tabular}
\end{table}

\section{Stage 1 Results}
\label{sec:stage1}

\subsection{P Dominates R Across All Scales}

Table~\ref{tab:stage1_best} summarizes the best config per scale and key trends from all 64 Stage 1 configurations.

\begin{table}[htbp]
\centering
\caption{Best Stage 1 config per scale (step 3{,}000). $\varrho$: mean spectral radius.}
\label{tab:stage1_best}
\footnotesize
\begin{tabular}{ccccccc}
\toprule
$d$ & $R^*$ & $P^*$ & \pplwiki & \ppltiny & \ppledu & $\varrho$ \\
\midrule
256 & 6 & 6 & 184.9 & 12.31 & 161.3 & 0.8927 \\
512 & 8 & 6 & 136.4 & 8.40  & 114.5 & 0.8936 \\
768 & 8 & 6 & 115.0 & 7.06  & 95.4  & 0.8956 \\
1024 & 8 & 6 & 97.73 & 6.04  & 82.0  & 0.8966 \\
\bottomrule
\end{tabular}
\end{table}

In every case, $P=6$ is best. The $P$ ordering ($P=6 > P=4 > P=3 > P=2$) holds without exception across all $R$ values and all four scales.

\subsection{R Benefit at Stage 1 vs.\ Stage 2}

\begin{figure}[htbp]
  \centering
  \includegraphics[width=\linewidth]{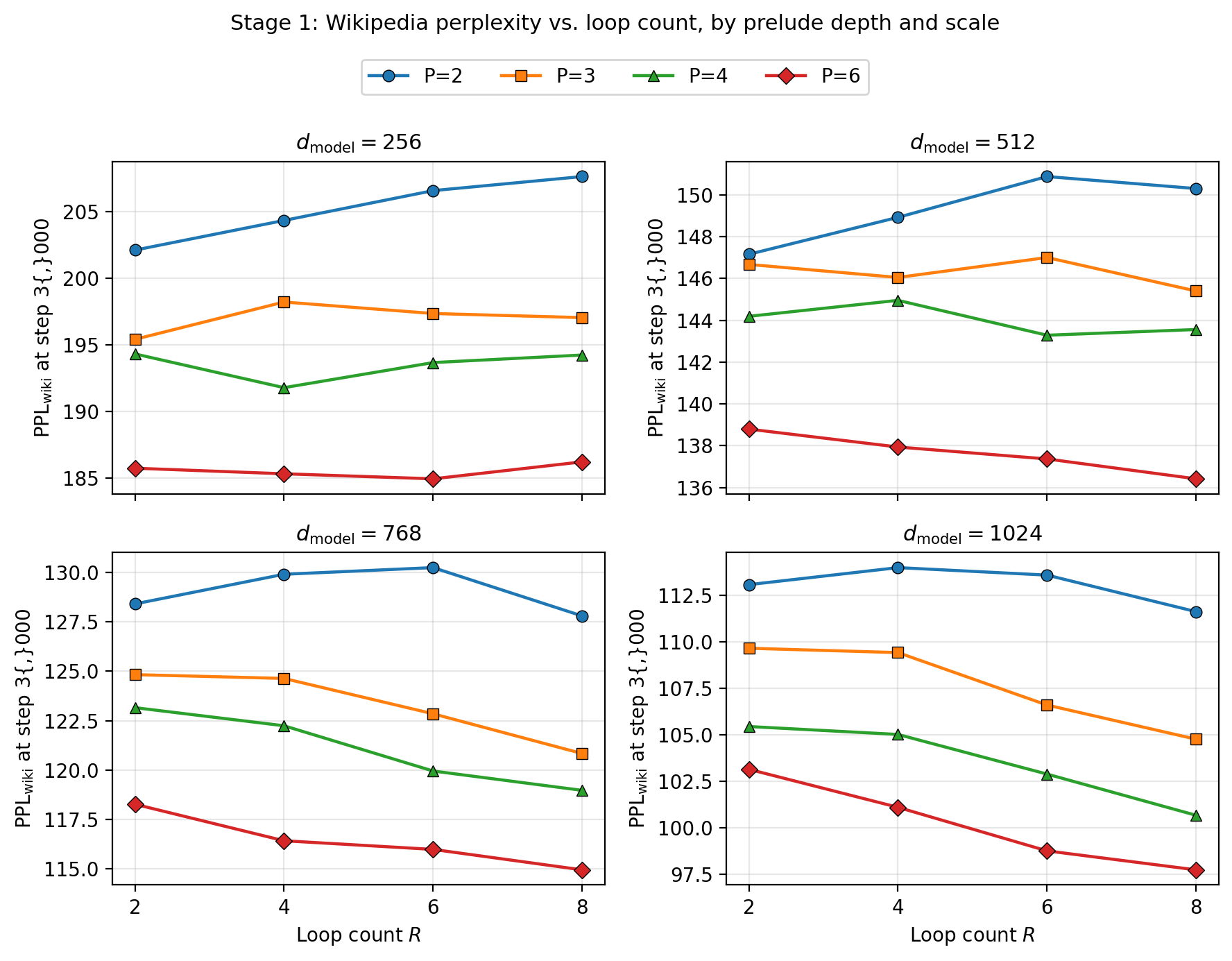}
  \caption{Stage 1 \pplwiki{} as a function of loop count $R$, with one panel per scale and one curve per prelude depth $P$ (step 3{,}000). The Stage 1 picture suggests $R$-benefit grows monotonically with $d$. Stage 2 results at the longer training budget reverse this trend; see Table~\ref{tab:r_benefit_s2}.}
  \label{fig:ppl_vs_r}
\end{figure}

At the Stage 1 training budget of 3{,}000 steps ($\sim 100$M tokens), the benefit of additional recurrence depth appeared to grow monotonically with model width: from a slight regression of $-0.25\%$ at $d{=}256$ to a $+5.24\%$ improvement at $d{=}1024$ on \pplwiki{} (Table~\ref{tab:r_benefit_s1}). Figure~\ref{fig:ppl_vs_r} visualizes this trend across $R \in \{2, 4, 6, 8\}$. The Stage 1 conclusion was that recurrence depth and representational width are complementary: larger models extract more from each additional loop because their prelude output is richer.

\begin{table}[htbp]
\centering
\caption{Stage 1 (step 3{,}000) $R$-benefit at $P{=}6$: \pplwiki{} improvement from $R{=}2$ to $R{=}8$.}
\label{tab:r_benefit_s1}
\footnotesize
\begin{tabular}{cccccr}
\toprule
$d$ & $R{=}2$ & $R{=}4$ & $R{=}6$ & $R{=}8$ & Gain \\
\midrule
256  & 185.7 & 185.3 & 184.9 & 186.2 & $-0.25\%$ \\
512  & 138.8 & 137.9 & 137.4 & 136.4 & $+1.72\%$ \\
768  & 118.3 & 116.4 & 116.0 & 115.0 & $+2.81\%$ \\
1024 & 103.1 & 101.1 & 98.75 & 97.73 & $+5.24\%$ \\
\bottomrule
\end{tabular}
\end{table}

At the Stage 2 training budget of 30{,}500 steps ($\sim 1$B tokens, $\sim 10\times$ Stage 1), the trend reverses. Table~\ref{tab:r_benefit_s2} reports 3-seed mean \pplwiki{} across $R \in \{6, 8, 10\}$ at every scale. $R$-benefit is positive only at $d{=}256$, where $R{=}8$ improves on $R{=}6$ by $2.1\%$. At every $d \geq 512$, $R{=}6$ is the best of the three on all three perplexity metrics; going from $R{=}6$ to $R{=}10$ at $d{=}1024$ yields a $0.7\%$ \emph{regression}. The pattern is consistent across \ppltiny, \pplwiki, and \ppledu.

\begin{table}[htbp]
\centering
\caption{Stage 2 (step 30{,}500) $R$-benefit at $P{=}6$: 3-seed mean \pplwiki{} across $R \in \{6, 8, 10\}$. Best $R$ per scale is bolded. Gain reports $R{=}6 \to R{=}10$.}
\label{tab:r_benefit_s2}
\footnotesize
\begin{tabular}{ccccr}
\toprule
$d$ & $R{=}6$ & $R{=}8$ & $R{=}10$ & Gain $R{=}6 \to R{=}10$ \\
\midrule
256  & 41.165 & \textbf{40.311} & 40.714 & $+1.10\%$ \\
512  & \textbf{27.097} & 27.203 & 27.378 & $-1.04\%$ \\
768  & \textbf{22.850} & 22.863 & 22.961 & $-0.48\%$ \\
1024 & \textbf{20.354} & 20.402 & 20.495 & $-0.69\%$ \\
\bottomrule
\end{tabular}
\end{table}

The Stage 1 prediction that $R$-benefit grows with $d$ is therefore not supported by Stage 2 measurements at this training budget; the direction is reversed. Three explanations are consistent with the data: (i) at full training, the prelude has absorbed enough signal that additional loops over a frozen anchor provide little new information, regardless of $d$; (ii) the LTI gate settles in an aggressive-discard regime (Section~\ref{sec:spectral}: $\varrho^{10} \approx 0.16$ at $d{=}1024$, $R{=}10$), so additional iterations contribute proportionally less; (iii) the training corpus (next-token prediction on natural prose with no code, math, or formal reasoning) may not contain enough examples that reward iterative refinement, so the recurrent core has no incentive to develop the per-iteration utility that would benefit higher $R$ at scale. Distinguishing (i)/(ii) from (iii) requires either training to fuller convergence (the extended-training experiment in Section~\ref{sec:discussion}) or training on a more iteration-reward-bearing corpus, both left for follow-up work.

\subsection{Scale Dominates Hyperparameters}

The weakest $d=1024$ config ($R=2$, $P=2$, \pplwiki{} $= 113.1$) outperforms the \emph{best} $d=768$ config ($R=8$, $P=6$, \pplwiki{} $= 115.0$). Within a scale, $P$ and $R$ matter; across scales, $d_\text{model}$ dominates. This motivates the Stage 2 framing: given a fixed parameter budget, does \cart{}'s architecture extract more quality per parameter than a Dense transformer?

\subsection{Loss Curves and Training Dynamics}

\begin{figure}[htbp]
  \centering
  \includegraphics[width=\columnwidth]{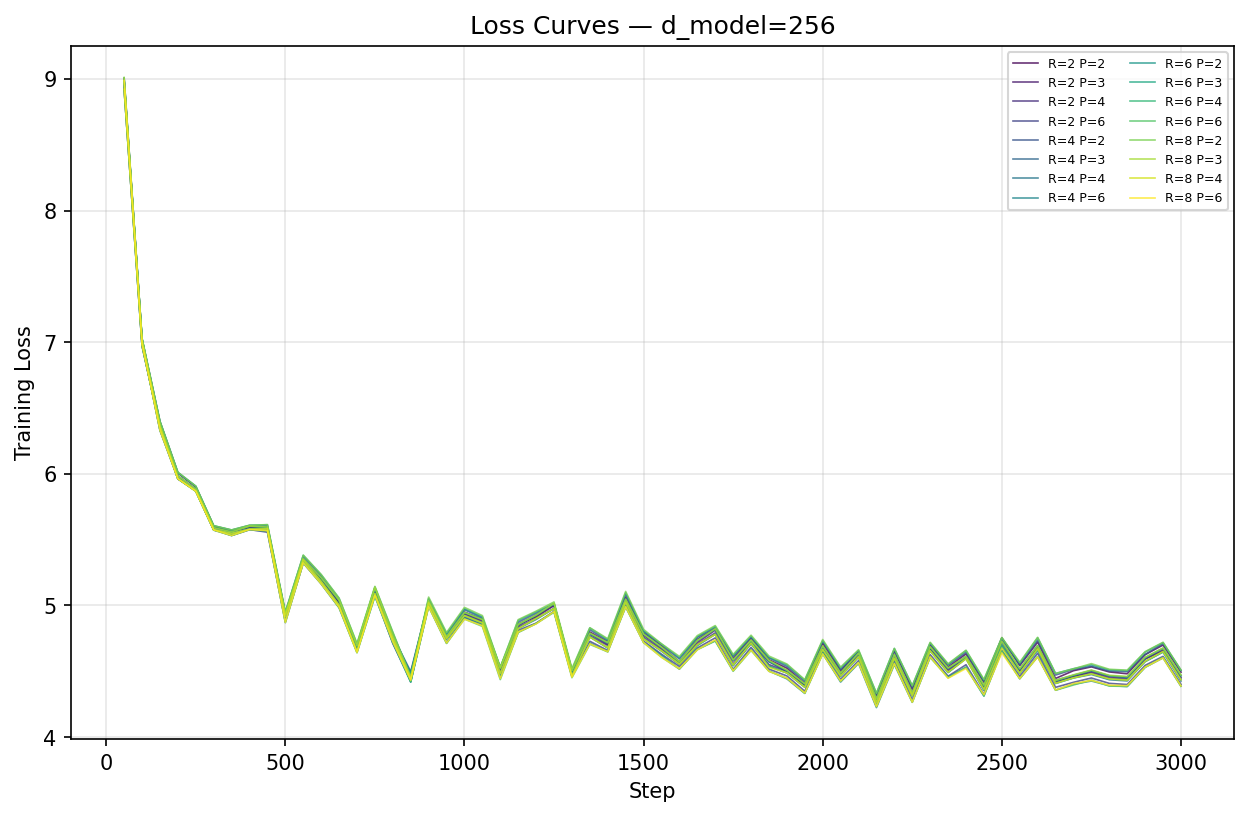}
  \caption{Training loss curves for all $d=256$ Stage 1 configs. All configs are still slowly declining at step 3{,}000; no plateau is reached. Stage 1 is a relative ranking, not a convergence measurement.}
  \label{fig:loss256}
\end{figure}

Figure~\ref{fig:loss256} shows training curves for $d=256$. All curves are still declining at step 3{,}000, confirming Stage 1 measures relative quality, not convergence.

\section{Spectral Radius Analysis}
\label{sec:spectral}

After full Stage 2 training (30{,}500 steps, $\sim$1B tokens), the LTI gate's spectral radius $\varrho$ settles in a narrow band $[0.79, 0.83]$, depending on both $d$ and $R$. Table~\ref{tab:rho} reports 3-seed mean $\varrho$ at step 30{,}500 across all 36 runs.

\begin{table}[htbp]
\centering
\caption{Final spectral radius $\varrho$ (3-seed mean at step 30{,}500) across all 36 Stage 2 configurations, $P{=}6$.}
\label{tab:rho}
\footnotesize
\begin{tabular}{ccccc}
\toprule
$d$ & $R{=}6$ & $R{=}8$ & $R{=}10$ & $\Delta$ from $d{=}256$ (R=6) \\
\midrule
256  & 0.790 & 0.795 & 0.799 & --- \\
512  & 0.800 & 0.804 & 0.807 & $+0.010$ \\
768  & 0.814 & 0.816 & 0.820 & $+0.024$ \\
1024 & 0.823 & 0.826 & 0.830 & $+0.033$ \\
\bottomrule
\end{tabular}
\end{table}

\begin{figure}[htbp]
  \centering
  \includegraphics[width=\columnwidth]{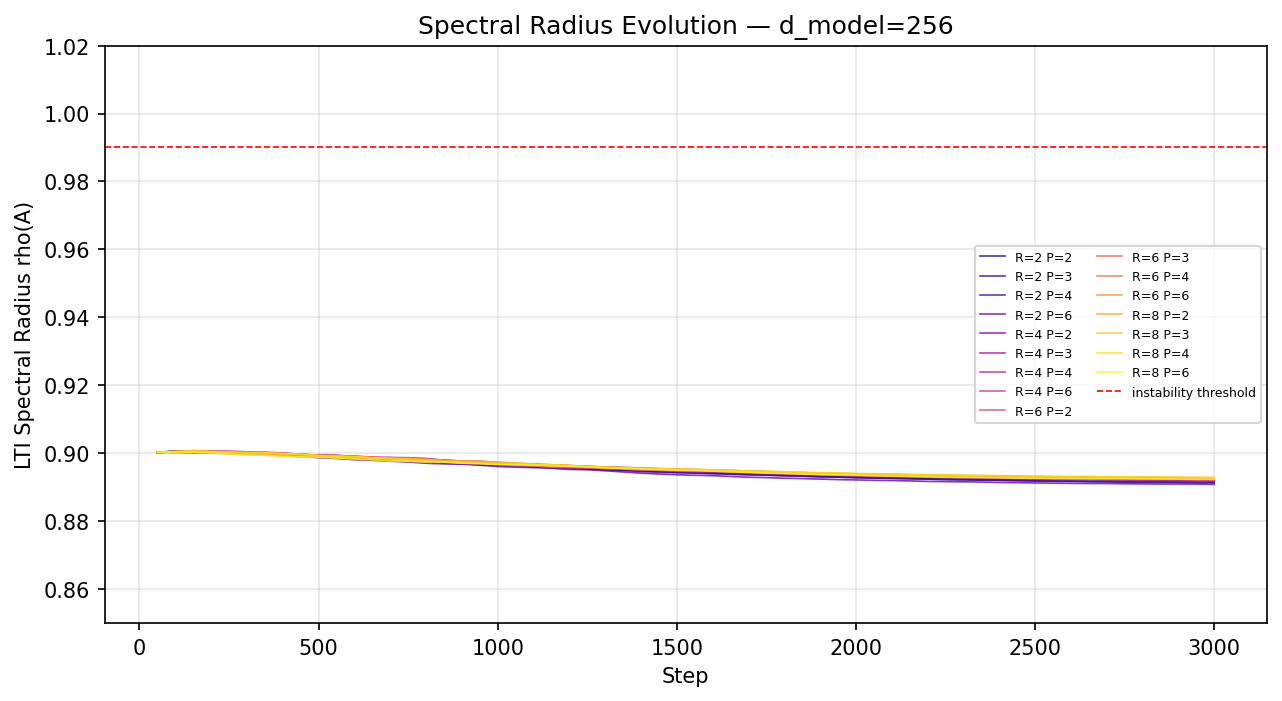}
  \caption{Stage 1 (3{,}000 step) spectral radius $\varrho$ trajectories for all $d{=}256$ configurations. The Stage 1 clustering near $\varrho \approx 0.893$ visible here is a mid-training transient; Stage 2 final values settle 0.09 to 0.10 lower (Table~\ref{tab:rho}).}
  \label{fig:rho256}
\end{figure}

\paragraph{Reconciling with Stage 1.} Stage 1 reported $\varrho \approx 0.893$ as a ``universal'' value across all 64 configurations, measured at step 3{,}000 (Section~\ref{sec:stage1}). Figure~\ref{fig:rho256} illustrates the Stage 1 trajectories at $d{=}256$, which do cluster near $0.893$ at the 3{,}000-step mark. Stage 2 reveals that this was a transient: $\varrho$ continues to decay through full training and only settles much later. For $d{=}1024$, $R{=}6$, seed 42, the trajectory runs $\varrho \approx 0.899$ at step 500, $0.886$ at step 5{,}000, $0.860$ at step 15{,}000, $0.834$ at step 25{,}000, and $0.828$ at step 30{,}500. The Stage 1 ``$0.5^{1/R}$ coincidence'' at $R{=}6$ ($\varrho \approx 0.8909$) does not survive to full training; final values are $\sim 0.06$ to $0.10$ below that line.

\paragraph{Scale-dependent retention.}
Mean $\varrho$ rises monotonically with $d$ at every fixed $R$, gaining $+0.033$ from $d{=}256$ to $d{=}1024$ at $R{=}6$. Larger models prefer slower per-iteration decay: a richer context anchor $e$ encodes more, and a higher $\varrho$ retains more of the previous hidden state across each gate application. The directional trend matches the Stage 1 observation, but the magnitude is substantially larger ($+0.033$ at full training vs.\ $+0.0042$ at 3{,}000 steps) and is no longer dominated by measurement noise.

\paragraph{Weak R-dependence (new).}
At every $d$, $\varrho$ also rises with $R$: $+0.008$ to $+0.009$ from $R{=}6$ to $R{=}10$. Stage 1 missed this because $R$-dependence is small compared to the mid-training transient. The interpretation is that when more iterations are available, the gate retains slightly more per-iteration state, but only slightly: about $1\%$ relative change per pair of additional loops. The model treats ``more loops'' as a mild reason to retain more, not a strong reason.

\paragraph{Residual persistence.}
The fraction of the initial residual that survives all $R$ iterations is $\varrho^R$. At $d{=}1024$, $R{=}6$: $\varrho^6 = 0.823^6 \approx 0.311$. At $d{=}1024$, $R{=}10$: $\varrho^{10} = 0.830^{10} \approx 0.155$. The gate is in an aggressive-discard regime in both cases, throwing away $69$--$85\%$ of the prelude-driven residual through the loop. This is consistent with the absence of R-benefit reported in Section~\ref{sec:stage2}: a gate that discards most of each iteration's output cannot strongly differentiate ``run for 6 iterations'' from ``run for 10 iterations.'' The mechanism the LTI gate has learned at this training budget is closer to ``use the prelude representation, refine modestly, then commit'' than to ``integrate over many refinement steps.''

\section{Stage 2 Results}
\label{sec:stage2}

\subsection{Stage 2 Final Results}

Table~\ref{tab:stage2_results} reports mean perplexity across 3 seeds for all 36 Stage 2 configurations plus the $P{=}4$ depth-ablation spot-check at $d{=}1024, R{=}10$ (Section~\ref{sec:hyp_vs_outcome}).

\begin{table*}[t]
\centering
\caption{Stage 2 results: mean perplexity across 3 seeds (30{,}500 steps, $\sim$1B tokens, seq\_len=1024). Per-seed standard deviations are summarized in Section~\ref{sec:repro}. Dense baseline at $d{=}1024$: 7-layer dense transformer, parameter-matched to CART on stored parameters (3-seed mean). Smaller-scale Dense baselines were not pursued in this study. A 12-layer Dense variant matched to CART's effective parameter count appears in Table~\ref{tab:stage2_dense12l}. Lower is better.}
\label{tab:stage2_results}
\footnotesize
\begin{tabular}{cccccccccc}
\toprule
\multirow{2}{*}{$d$} & \multirow{2}{*}{$R$} & \multirow{2}{*}{$P$} & \multicolumn{3}{c}{Mean CART} & \multicolumn{3}{c}{Dense Baseline} & \multirow{2}{*}{Status} \\
\cmidrule(lr){4-6}\cmidrule(lr){7-9}
& & & \ppltiny & \pplwiki & \ppledu & \ppltiny & \pplwiki & \ppledu & \\
\midrule
256 & 6  & 6 & 4.343 & 41.17 & 41.52 & --- & --- & --- & Complete \\
256 & 8  & 6 & 4.343 & 40.31 & 40.77 & --- & --- & --- & Complete \\
256 & 10 & 6 & 4.371 & 40.71 & 41.22 & --- & --- & --- & Complete \\
\midrule
512 & 6  & 6 & 3.360 & 27.10 & 27.44 & --- & --- & --- & Complete \\
512 & 8  & 6 & 3.366 & 27.20 & 27.60 & --- & --- & --- & Complete \\
512 & 10 & 6 & 3.367 & 27.38 & 27.55 & --- & --- & --- & Complete \\
\midrule
768 & 6  & 6 & 3.010 & 22.85 & 23.15 & --- & --- & --- & Complete \\
768 & 8  & 6 & 3.023 & 22.86 & 23.30 & --- & --- & --- & Complete \\
768 & 10 & 6 & 3.024 & 22.96 & 23.38 & --- & --- & --- & Complete \\
\midrule
1024 & 6  & 6 & 2.798 & 20.35 & 20.67 & \multirow{3}{*}{2.746} & \multirow{3}{*}{20.13} & \multirow{3}{*}{20.26} & Complete \\
1024 & 8  & 6 & 2.808 & 20.40 & 20.79 &  &  &  & Complete \\
1024 & 10 & 6 & 2.811 & 20.50 & 20.81 &  &  &  & Complete \\
\midrule
1024 & 10 & 4 & 2.899 & 21.42 & 21.72 & --- & --- & --- & Reference \\
\bottomrule
\end{tabular}
\end{table*}

The final row reports $P{=}4$ as a depth-ablation spot-check at the largest scale. The active Stage 2 sweep was $P{=}6$ throughout; $P{=}4$ was run only at $d{=}1024, R{=}10$ (3 seeds) to bound the cost of reducing prelude depth at the high end. Dropping from $P{=}6$ to $P{=}4$ costs $+0.088$ on \ppltiny{}, $+0.92$ on \pplwiki{}, and $+0.91$ on \ppledu{} ($\sim 4.5\%$ on the natural-language sets).

The 7L Dense baseline in Table~\ref{tab:stage2_results} is parameter-matched to CART on \emph{stored} parameters (Dense 7L $\approx$ CART d=1024). CART's recurrent core also accumulates \emph{effective} parameters via the $R$-iteration leverage (105M at $R{=}6$, see Table~\ref{tab:chinchilla}); to test the effective-parameter parity question separately, we trained a 12-layer Dense baseline at $d{=}1024$ matched to that count. Table~\ref{tab:stage2_dense12l} reports the result and the deltas vs the best CART config ($P{=}6, R{=}6$).

\begin{table}[htbp]
\centering
\caption{Dense 12L (12-layer dense transformer) at $d{=}1024$, parameter-matched to CART's effective parameter count at $R{=}6$ ($\sim 105$M). 3-seed mean $\pm$ std. Delta $\Delta$ is Dense $-$ CART; negative means Dense is better.}
\label{tab:stage2_dense12l}
\footnotesize
\begin{tabular}{lccc}
\toprule
 & \ppltiny & \pplwiki & \ppledu \\
\midrule
Dense 12L (3-seed mean)        & $2.592 \pm 0.003$ & $18.32 \pm 0.05$ & $18.50 \pm 0.06$ \\
CART $P{=}6, R{=}6$ (3-seed mean) & $2.798 \pm 0.013$ & $20.35 \pm 0.08$ & $20.67 \pm 0.14$ \\
\midrule
$\Delta$ (Dense $-$ CART)      & $-0.205$ ($-7.3\%$) & $-2.02$ ($-9.9\%$) & $-2.16$ ($-10.5\%$) \\
\bottomrule
\end{tabular}
\end{table}

\subsection{Findings from Complete Configurations}

\paragraph{Low-end scales: $d{=}256$ and $d{=}512$.}
At the two smallest widths, \cart{}'s 3-seed mean \pplwiki{} is $40.3$--$41.2$ at $d{=}256$ and $27.1$--$27.4$ at $d{=}512$ across $R \in \{6, 8, 10\}$, with \ppledu{} tracking \pplwiki{} closely and \ppltiny{} an order of magnitude lower on the narrow-vocabulary set (Table~\ref{tab:stage2_results}). Parameter-matched Dense baselines were trained only at $d{=}1024$ (Section~\ref{sec:setup}), so these widths are reported without a baseline; the binding \cart{}--Dense comparison is at $d{=}1024$ (Section~\ref{sec:hyp_vs_outcome}).

\paragraph{R insensitivity at $d{=}512$ and above.}
At $d{=}512$, $R{=}6$ and $R{=}8$ produce statistically indistinguishable perplexity (difference $< 0.01$ on \ppltiny). This R-flatness pattern persists at $d{=}768$ and $d{=}1024$: in both cases $R{=}6$ is the best of $\{6, 8, 10\}$ on all three perplexity metrics, with the gap to $R{=}10$ within $1\%$ at $d{=}768$ and within $0.7\%$ at $d{=}1024$. The Stage 1 prediction that $R$-benefit would grow substantially at $d{=}768$ and $d{=}1024$ is reversed at the Stage 2 training budget. See Section~\ref{sec:stage1} (Table~\ref{tab:r_benefit_s2}) for the full Stage 1 vs.\ Stage 2 comparison.

\paragraph{Cross-tokenizer reference (bits per byte).}
For comparison with models trained under different tokenizers, we report bits per byte (BPB) computed as $\text{BPB} = \log_2(\text{ppl}) \cdot t/b$, where $t/b$ is the average tokens-per-byte ratio of the Llama-2 tokenizer on the relevant evaluation domain ($\approx 0.27$ for English natural-language text). At $d{=}512$, mean across $R \in \{6, 8, 10\}$ and 3 seeds: $\text{BPB}_\text{wiki} = 1.286$, $\text{BPB}_\text{edu} = 1.292$. At $d{=}256$, mean: $\text{BPB}_\text{wiki} = 1.444$, $\text{BPB}_\text{edu} = 1.448$.

For order-of-magnitude context, GPT-2 small (124M parameters)~\citep{radford2019gpt2} reports $\sim 1.05$--$1.20$ BPB on WikiText-103 after training on $\sim 8$B tokens; Pythia-160M~\citep{biderman2023pythia} reports $\sim 1.30$--$1.50$ BPB on Wikipedia subsets at $\sim 300$B tokens. \cart{}'s $d{=}512$ figures of $1.29$ on Wikipedia at $\sim 1$B tokens (effective parameter count $55$--$77$M) are in the same range as these references at much larger token budgets, though direct comparison requires identical evaluation corpora and is left for the downstream-benchmarks evaluation (Section~\ref{sec:benchmarks}). The perplexity-to-BPB conversion formula and tokenizer ratios are provided in the released code.

\subsection{Cross-Scale Scaling}
\label{sec:scaling}

Stage 1 perplexity at the best $(R, P)$ configuration per scale fits a power law in $d_\text{model}$:
\[
\text{PPL}_\text{wiki}^\text{stage1}(d) \;\approx\; 2371 \cdot d^{-0.458}
\]
fit on four data points $(d, R{=}8, P{=}6)$ at step 3{,}000 ($r^2 \approx 0.999$). Each doubling of $d$ reduces \pplwiki{} by approximately $28\%$. The exponent of $-0.458$ is at the steeper end of the range reported for standard transformer scaling laws~\citep{kaplan2020scaling,hoffmann2022chinchilla}, which typically fit exponents on $d$ alone in the range $-0.3$ to $-0.4$ at sub-Chinchilla token budgets.

At the full Stage 2 training budget of $\sim 1$B tokens, the cross-scale picture is available at all four scales as 3-seed $R{=}6$ means. We report $R{=}6$ across all scales for a clean cross-scale comparison: $R{=}6$ is the best $R$ at every scale $d \geq 512$ on all three metrics, and at $d{=}256$ is within $2.1\%$ of the $R{=}8$ best on the natural-language sets while tying on \ppltiny.

The Stage 2 single-step exponents on \pplwiki{} soften progressively with scale: $-0.60$ from $d{=}256 \to d{=}512$, $-0.42$ from $d{=}512 \to d{=}768$, and $-0.40$ from $d{=}768 \to d{=}1024$. A 4-point power-law fit on the $R{=}6$ means yields $\text{PPL}_\text{wiki}^\text{stage2}(d) \approx 683 \cdot d^{-0.51}$, steeper than the Stage 1 fit of $-0.46$ but with visible softening at the larger end. 3-seed agreement at $d{=}1024$, $R{=}6$ is tight (standard deviation on \pplwiki across seeds: $0.07$; on \ppltiny: $0.010$).

Two interpretations of the softening remain consistent with the data: (a) the cross-scale fit is approximately $-0.50$ with some variance at each transition, and the per-segment softening is sample noise; or (b) per-$d$ efficiency genuinely diminishes above $d{=}512$, signaling early saturation of the prelude representation under the $\sim 1$B-token training budget. Interpretation (b) is the more parsimonious read of three successively shallower exponents, but distinguishing the two would require either a wider-than-three-seed sample at each scale or the extended-training experiment (Section~\ref{sec:discussion}) to push $d{=}1024$ closer to effective-Chinchilla and observe whether its exponent stiffens.

\begin{table}[htbp]
\centering
\caption{Cross-scale \pplwiki{} at $R{=}6$, $P{=}6$ across Stage 1 (step 3{,}000) and Stage 2 (step 30{,}500). Stage 2 entries are 3-seed means. Implied exponent computed from the prior scale entry.}
\label{tab:scaling}
\footnotesize
\begin{tabular}{rrrl}
\toprule
$d$ & Stage 1 \pplwiki & Stage 2 \pplwiki & Implied exp.\ vs.\ prior $d$ \\
\midrule
 256 & 186.20 & 41.17 (mean of 3) & ---                   \\
 512 & 136.42 & 27.10 (mean of 3) & $-0.60$ ($\times 2$)   \\
 768 & 114.96 & 22.85 (mean of 3) & $-0.42$ ($\times 1.5$) \\
1024 &  97.73 & 20.35 (mean of 3) & $-0.40$ ($\times 1.33$) \\
\bottomrule
\end{tabular}
\end{table}

Parameter-matched Dense baselines across all four scales were not run in this study. Two questions therefore remain open and are left for follow-up work: (i) whether \cart{}'s scaling exponent in $d$ is steeper or shallower than Dense's at parameter parity; and (ii) per-parameter and per-FLOP knowledge density relative to Dense. The first is the central question for the parameter-efficiency claim and the cleanest direction for a follow-up paper.

\subsection{Inference-Time Depth Scaling}
\label{sec:inference_R}

\cart{}'s frozen-KV design admits a property not available to architectures that recompute $K, V$ each iteration: $R$ at inference time can differ from $R$ at training time without retraining. The frozen $K, V$ tensors do not depend on $R$; the core block's weights are shared; the LIE table extends to $R{=}16$ in our implementation; and the LTI gate's $\varrho < 1$ guarantee ensures convergence as $R \to \infty$.

We evaluate this property on the best $d{=}1024$ Stage 2 model. Since $R{=}6$ produced the lowest perplexity at $d{=}1024$ across all three metrics (Section~\ref{sec:scaling}), the model selected for this experiment was trained at $R{=}6$, and we run inference at $R \in \{2, 4, 6, 8, 10, 12, 14, 16\}$ with no fine-tuning between settings. The four benchmark tasks (HellaSwag, ARC-Challenge, LAMBADA, PIQA) are run end-to-end at every $R$. The sweep covers both directions around the trained $R$: $R < 6$ tests inference-time compute \emph{reduction} (fewer iterations than trained), $R > 6$ tests inference-time compute \emph{scaling} (more iterations than trained, the deterministic-latency analogue to the variable-compute scaling reported by~\citet{geiping2025latent}).

Table~\ref{tab:variable_r} reports the results. The picture is bilateral: $R{=}6$, the training $R$, is the best inference setting on the three metrics that carry signal (HellaSwag, LAMBADA, PIQA), and performance degrades on both sides. Below the trained $R$, the degradation is sharp: LAMBADA perplexity rises from $163.7$ at $R{=}6$ to $328.8$ at $R{=}4$ ($2\times$ worse) and $2632.6$ at $R{=}2$ ($16\times$ worse). Above the trained $R$, the degradation is gentler but monotonic: LAMBADA reaches $269.4$ at $R{=}16$ ($65\%$ worse than $R{=}6$); HellaSwag drops by $\sim 0.6$ points, PIQA by $\sim 1.9$ points. ARC-Challenge varies by less than $5$ points across all eight inference $R$ values but does not provide a clear signal: at every $R \in \{4, 6, 8, 10, 12, 14, 16\}$, accuracy lies below the $25\%$ random baseline for a 4-choice task, indicating the model is confidently wrong rather than uncertain, and the apparent $R{=}2$ peak at $25.43\%$ reflects the model essentially answering randomly rather than learning anything useful from two iterations.

The shape of the degradation profile is direct evidence that the recurrent core has converged to a specific fixed-point structure tied to its trained $R$: too few iterations leave the representation under-developed (especially visible on LAMBADA's long-range coreference), while too many iterations push the state past its trained convergence point. This is a clean negative result for test-time compute scaling in \cart{} under the present training recipe; the cheapest path to test-time compute scaling for \cart{} appears to be training-time variable-$R$ (as in~\citet{prairie2026parcae}) rather than post-hoc extrapolation. The negative finding is consistent with the absent R-benefit observed during training (Section~\ref{sec:stage1}): the model does not extract per-iteration improvement on this corpus regime, regardless of when those iterations happen.

\begin{table}[htbp]
\centering
\caption{Variable-$R$ inference on the best $d{=}1024$ Stage 2 model, trained at $R{=}6$. HellaSwag, ARC-C, and PIQA report normalized accuracy; LAMBADA reports accuracy with perplexity in parentheses. Best value per column among the rows that carry signal (HellaSwag, LAMBADA, PIQA) is bolded. ARC-C values below the 4-choice $25\%$ random baseline indicate the model is confidently wrong rather than uncertain; the apparent $R{=}2$ ``best'' is essentially random.}
\label{tab:variable_r}
\footnotesize
\begin{tabular}{rcccc}
\toprule
Inference $R$ & HellaSwag & ARC-C & LAMBADA acc (ppl) & PIQA \\
\midrule
 2         & 27.91 & 25.43 &  7.74 (2632.6) & 55.01 \\
 4         & 27.71 & 23.38 & 18.01 ( 328.8) & 56.64 \\
 6 (train) & \textbf{28.04} & 21.84 & \textbf{23.04} (\textbf{163.7}) & \textbf{58.11} \\
 8         & 27.90 & 21.33 & 22.12 ( 173.0) & 57.83 \\
10         & 27.97 & 21.59 & 21.41 ( 190.6) & 57.40 \\
12         & 27.61 & 21.93 & 20.65 ( 214.2) & 56.58 \\
14         & 27.47 & 21.84 & 19.46 ( 246.0) & 56.47 \\
16         & 27.42 & 21.93 & 18.55 ( 269.4) & 56.20 \\
\bottomrule
\end{tabular}
\end{table}

\subsection{Downstream Benchmarks}
\label{sec:benchmarks}

Perplexity establishes that \cart{} produces a calibrated language model, but downstream task accuracy is the harder test. We evaluate \cart{} at the best $R$ for each of the four scales (one model per scale, single seed) on four standard zero-shot benchmarks via the LM Evaluation Harness~\citep{eval-harness}: HellaSwag (commonsense completion), ARC-Challenge (grade-school science), LAMBADA (long-range coreference), and PIQA (physical commonsense). The wikitext perplexity task is excluded due to a known harness hang.

\begin{table}[htbp]
\centering
\caption{Zero-shot downstream benchmarks for \cart{} at best $R$ per scale (one seed). HellaSwag and ARC-Challenge report normalized accuracy ($\mathrm{acc\_norm}$); LAMBADA reports accuracy with perplexity in parentheses; PIQA reports raw accuracy. Best $R$ at each scale is selected by lowest mean \pplwiki{}. Parameter-matched Dense baselines were not run in this study.}
\label{tab:benchmarks}
\footnotesize
\begin{tabular}{rcccccc}
\toprule
$d$ & Best $R$ & HellaSwag & ARC-C & LAMBADA acc (ppl) & PIQA & Dense match? \\
\midrule
 256 &  8 & 26.49 & 21.59 & 8.66 (2079) & 54.79 & --- \\
 512 &  6 & 27.10 & 22.87 & 16.15 (507) & 55.60 & --- \\
 768 &  6 & 27.05 & 22.87 & 20.42 (251) & 57.67 & --- \\
1024 &  6 & 28.04 & 21.84 & 23.02 (164) & 58.11 & --- \\
\bottomrule
\end{tabular}
\end{table}

These four benchmarks cover commonsense reasoning (HellaSwag), scientific factual recall (ARC-C), long-range linguistic coherence (LAMBADA), and physical-world plausibility (PIQA). The cross-scale data reveal a divided pattern at sub-1B parameter counts. LAMBADA and PIQA scale smoothly across all four widths: LAMBADA accuracy rises monotonically from $8.66\%$ at $d{=}256$ through $16.15\%$, $20.42\%$, to $23.02\%$ at $d{=}1024$, with perplexity halving at every step ($2079 \to 507 \to 251 \to 164$); PIQA accuracy rises monotonically from $54.79\%$ to $58.11\%$. HellaSwag and ARC-C, by contrast, show a near-plateau across $d{=}512$, $768$, and $1024$ ($27.10$, $27.05$, $28.04$ on HellaSwag; $22.87$, $22.87$, $21.84$ on ARC-C, with the $d{=}1024$ ARC-C value within seed noise of random at $25\%$ on the 4-choice task). These benchmarks test capabilities that are widely reported to emerge at scale rather than improving smoothly, and our results are consistent with the model remaining below the capability threshold for those tasks at the parameter and training-token budgets we tested.

For reference, GPT-2 small (124M parameters, $\sim 8$B training tokens) reports approximately $29\%$ HellaSwag, $22\%$ ARC-C, $46\%$ LAMBADA accuracy, and $62\%$ PIQA. CART at $d{=}1024$ ($\sim 179$M effective parameters, $1$B training tokens) reaches $28.04\%$ HellaSwag (within $1$ point of GPT-2 small) and $21.84\%$ ARC-C (matching within noise), at one-eighth the training data. LAMBADA accuracy remains substantially below GPT-2 small's, which we attribute primarily to the training-token gap; the extended-training experiment (Section~\ref{sec:discussion}) is the cleaner test for whether additional training on the same architecture closes that gap.

\paragraph{Few-shot evaluation at $d{=}1024$.}
In addition to zero-shot scores, we report 5-shot accuracy on the best $d{=}1024$ model on the same four tasks (Table~\ref{tab:fewshot}). 5-shot evaluation tests whether the model can use in-context examples to improve task accuracy, an emergent capability that is typically undetectable at sub-100M parameter counts but begins to appear at $\sim 200$M effective parameters. We omit few-shot evaluation at $d \in \{256, 512, 768\}$ because zero-shot accuracy at those scales is already near random, leaving no measurable headroom for in-context examples to help.

The results are mixed and informative. On ARC-Challenge, 5-shot improves over 0-shot by $1.2$ points ($21.84 \to 23.04$, both still below the $25\%$ 4-choice random baseline); on HellaSwag, by $0.3$ points ($28.04 \to 28.36$). These small but positive deltas indicate the model does extract some useful signal from in-context demonstrations on multiple-choice tasks. On LAMBADA, by contrast, 5-shot accuracy drops sharply ($23.02 \to 15.25$, a $-7.8$ point regression) and perplexity more than doubles ($163.7 \to 366.3$). PIQA also regresses modestly ($-2.0$ points). The LAMBADA regression is consistent with format mismatch: the task expects a single contiguous passage with a missing final word, but the 5-shot evaluation prepends five demonstration passages that shift the input out of the model's trained sequence-length and structural distribution. PIQA's small regression has a similar explanation: the model's representation of physical-commonsense questions does not readily generalize to a five-times-prepended input. We interpret the asymmetry as evidence that emergent few-shot capability is only partially present at this parameter and training-token budget: the model can leverage demonstrations when the format is amenable but cannot yet robustly use them under format shifts.

\begin{table}[htbp]
\centering
\caption{Zero-shot vs.\ 5-shot accuracy at $d{=}1024$, $R{=}6$, seed 271 (the best $d{=}1024$ Stage 2 model). HellaSwag, ARC-C, and PIQA report normalized accuracy. LAMBADA reports accuracy with perplexity in parentheses.}
\label{tab:fewshot}
\footnotesize
\begin{tabular}{lcccc}
\toprule
Shots & HellaSwag & ARC-C & LAMBADA acc (ppl) & PIQA \\
\midrule
0-shot & 28.04 & 21.84 & 23.02 (163.7) & 58.11 \\
5-shot & 28.36 & 23.04 & 15.25 (366.3) & 56.09 \\
\bottomrule
\end{tabular}
\end{table}

\subsection{Summary: Hypothesis vs.\ Outcome}
\label{sec:hyp_vs_outcome}

\cart{}'s architectural bet was that shared-weight recurrence anchored on a once-computed prelude representation would extract additional capacity per stored parameter. We summarize what the Stage 2 results say about that bet.

\paragraph{Predicted.}
(1) At fixed stored parameters, recurrence increases effective capacity relative to a parameter-matched Dense transformer; the leverage ratio in Table~\ref{tab:chinchilla} converts stored parameters into effective capacity by a factor of $\sim R/2$ to $R$ depending on prelude depth. (2) The benefit of recurrence depth grows with model width $d$, predicted by the Stage 1 monotonic pattern from $-0.25\%$ at $d{=}256$ to $+5.24\%$ at $d{=}1024$ ($R{=}2 \to R{=}8$ on \pplwiki, Table~\ref{tab:r_benefit_s1}). (3) Once-computed frozen $K, V$ from the prelude provides a clean efficiency gain (skipping $2R$ projections per token) without expressive cost; the prelude representation is rich enough that iterative refinement of queries against fixed keys suffices.

\paragraph{Observed at $d{=}1024$, full Stage 2 budget.}
(1) Dense 7L (parameter-matched on stored parameters, 3-seed mean) beats \cart{} $P{=}6, R{=}6$ by $1.1\%$ on \pplwiki, $2.0\%$ on \ppledu, and $1.8\%$ on \ppltiny{} (Tables~\ref{tab:stage2_results} and~\ref{tab:stage2_dense12l}). The seed agreement of both models is tight enough that the gap is signal, not noise (Section~\ref{sec:repro}). (2) Dense 12L (parameter-matched on effective parameters, 3-seed mean) beats \cart{} by $9.9\%$ on \pplwiki and $10.5\%$ on \ppledu (Table~\ref{tab:stage2_dense12l}). (3) $R$-benefit is absent: $R{=}6$ is the best of $\{6, 8, 10\}$ at every $d \geq 512$, with the gap to $R{=}10$ within $1\%$ (Table~\ref{tab:r_benefit_s2}). The Stage 1 prediction of growing $R$-benefit at large $d$ is reversed at the Stage 2 budget. (4) Prelude depth dominates loop count within the spot-check we ran: $P{=}4$ at $d{=}1024, R{=}10$ costs $+0.92$ on \pplwiki, larger than the gap from any $R$ choice at the same scale (Table~\ref{tab:stage2_results}).

\paragraph{What this means.}
The parameter-efficiency thesis fails at the binding test ($d{=}1024$). The small ($\sim 1$--$2\%$) stored-parameter gap is consistent with \cart{} being competitive but not advantageous on a per-stored-parameter basis at this scale. The larger ($\sim 10\%$) effective-parameter gap is the stronger negative: the shared-weight leverage was supposed to convert into effective capacity, and the data shows it does not under our training recipe at $\sim 1$B tokens. Two architectural choices in \cart{} (the frozen $K, V$ anchor in the recurrent core, and the single shared-weight core looped $R$ times) are the most direct candidates for the bottleneck. Section~\ref{sec:ablations} reports diagnostic ablations testing each.

\section{Diagnostic Ablations}
\label{sec:ablations}

The $d{=}1024$ result (Section~\ref{sec:hyp_vs_outcome}) raises a focused question: which of \cart{}'s architectural choices is responsible for the gap with Dense? We report a sequence of single-seed ablations against the candidate bottlenecks: frozen $K, V$ from the prelude, useful contribution from loop iterations beyond the first, shared identical weights across all $R$ iterations, the cross-attention-against-anchor structure, the HyperConnection blending pattern, and the LTI sigmoid gate. Each ablation trains a variant at $d{=}1024, R{=}6, P{=}6, \text{seed}{=}42$ (the diagnostic corner where the Dense gap is cleanest) with all other hyperparameters identical to the baseline; the same seed throughout makes each $\Delta$ vs baseline isolate the architectural change. Statistical caveats apply: with one seed per ablation versus three for the baseline, only effects exceeding the baseline's seed-noise band ($\sim 0.08$ on \pplwiki, $\sim 2\sigma \approx 0.16$) are interpretable as signal; smaller effects are reported as null but acknowledged as within seed noise.

\subsection{Ablation X: Unfreezing the KV Cache}
\label{sec:abl_kv}

\paragraph{Hypothesis.}
Frozen $K, V$ may be the architectural ceiling. The central architectural distinction between \cart{} and Hyperloop~\citep{zeitoun2026hyperloop} is the treatment of the key-value representations in the recurrent core. Hyperloop recomputes $K, V$ from the evolving hidden state at each iteration, effectively running $R$ full self-attention layers with shared weights. \cart{} computes $K, V$ once from the prelude and freezes them for all $R$ iterations. The frozen design's appealing properties (lower per-loop FLOPs, a clean separation between context encoding and iterative refinement) may come at the cost of capacity: every iteration cross-attends against the same prelude-derived representation regardless of how the recurrent state evolves.

\paragraph{Intervention.}
At each loop iteration $r$, recompute $K, V = \texttt{kv\_proj}(h_\text{input})$ from the current hidden state instead of using the cached pre-loop $K, V = \texttt{kv\_proj}(e)$. The same \texttt{kv\_proj} weights are reused; no new parameters are added. Stored parameter count is unchanged; effective parameters increase by $\sim 1.9\%$ at $R{=}6$ because \texttt{kv\_proj} is now called $R$ times per forward pass instead of once.

\paragraph{Result.}

\begin{table}[htbp]
\centering
\caption{Unfreezing $K, V$ at $d{=}1024, R{=}6, P{=}6$, seed 42. Baseline is the 3-seed mean \cart{} run from Table~\ref{tab:stage2_results}. Lower is better.}
\label{tab:abl_unfreezekv}
\footnotesize
\begin{tabular}{lcccc}
\toprule
 & \ppltiny & \pplwiki & \ppledu & seeds \\
\midrule
\cart{} baseline ($K, V$ frozen)         & $2.798 \pm 0.013$ & $20.35 \pm 0.08$ & $20.67 \pm 0.14$ & 3 \\
\cart{} unfrozen $K, V$                  & $2.783$           & $20.14$          & $20.52$          & 1 \\
\midrule
$\Delta$ (unfrozen $-$ baseline)         & $-0.015$ ($-0.5\%$) & $-0.21$ ($-1.0\%$) & $-0.15$ ($-0.7\%$) & --- \\
\bottomrule
\end{tabular}
\end{table}

The unfrozen-$K, V$ variant remained within the baseline's own seed-noise band on every metric ($1\sigma$ on \pplwiki: $0.08$; observed shift: $0.21$, roughly $2.6\sigma$ but with $n{=}1$ on the ablation side). Train loss curves tracked the baseline within $\pm 0.05$ nats across all 610 logged steps from step 50 to step 30{,}500. Despite the architectural change, \cart{} remains slightly behind Dense 7L ($20.13$ \pplwiki) and substantially behind Dense 12L ($18.32$ \pplwiki, Table~\ref{tab:stage2_dense12l}).

\paragraph{Interpretation.}
Frozen $K, V$ is not the architectural choice responsible for the gap with Dense at $d{=}1024$. The original \cart{} design's compute saving ($2R$ projections per token skipped per sequence) and the architectural simplification it provides remain real benefits, but the hypothesis that allowing $K, V$ to evolve would unlock additional capacity does not hold empirically at this scale and training budget. The bottleneck must be elsewhere in the architecture.

\subsection{Ablation Y: Recurrence-Depth Reduction ($R{=}1$)}
\label{sec:abl_r1}

\paragraph{Hypothesis.}
The recurrent core may not be extracting useful capacity from iterations beyond the first. If so, removing $R{-}1$ of the iterations should leave perplexity essentially unchanged, and \cart{}'s "recurrence" framing reduces to architectural decoration.

\paragraph{Intervention.}
Train \cart{} with $n_\text{loops} = 1$ at $d{=}1024, P{=}6, \text{seed}{=}42$. The for-loop in \texttt{CART.forward} runs once; HyperConnection, LIE, and the LTI gate all degenerate gracefully (LIE emits \texttt{pe[0]}, the ring buffer initializes to three copies of $e$ which combine to $e$, the LTI gate applies once). Effective parameters at $R{=}1$ equal stored parameters ($\sim 75$M), parameter-matching Dense 7L by construction. The single-iteration core does the same cross-attention against the frozen prelude-derived $K, V$, so this isolates "does the loop contribute anything?" without changing any other architectural detail.

\paragraph{Result.}

\begin{table}[htbp]
\centering
\caption{Recurrence-depth ablation at $d{=}1024, P{=}6$, seed 42. At $R{=}1$, effective $=$ stored ($\sim 75$M), parameter-matching Dense 7L. Lower is better.}
\label{tab:abl_r1}
\footnotesize
\begin{tabular}{lcccc}
\toprule
 & \ppltiny & \pplwiki & \ppledu & seeds \\
\midrule
\cart{} baseline ($R{=}6$, $\sim 75$M stored)  & $2.798 \pm 0.013$ & $20.35 \pm 0.08$ & $20.67 \pm 0.14$ & 3 \\
\cart{} $R{=}1$ ($\sim 75$M, no leverage)      & $2.811$           & $20.57$          & $20.77$          & 1 \\
Dense 7L ($\sim 75$M)                          & $2.746 \pm 0.002$ & $20.13 \pm 0.06$ & $20.26 \pm 0.04$ & 3 \\
\midrule
$\Delta$ ($R{=}1$ $-$ baseline $R{=}6$)        & $+0.013$ ($+0.5\%$) & $+0.22$ ($+1.1\%$) & $+0.10$ ($+0.5\%$) & --- \\
$\Delta$ ($R{=}1$ $-$ Dense 7L)                & $+0.065$ ($+2.4\%$) & $+0.44$ ($+2.2\%$) & $+0.51$ ($+2.5\%$) & --- \\
\bottomrule
\end{tabular}
\end{table}

The $R{=}1$ variant is slightly worse than the $R{=}6$ baseline on every metric, by $0.5$--$1.1\%$. The \pplwiki gap of $+0.22$ is approximately $2.7\sigma$ of the baseline's seed-noise band, with $n{=}1$ on the ablation side, putting it at the edge of interpretable signal. Train loss curves track the baseline closely throughout, with the $R{=}6$ baseline pulling marginally ahead in the second half of training.

At parameter parity with Dense 7L (since $R{=}1$ has no leverage multiplier), \cart{} $R{=}1$ loses by $2.2$--$2.5\%$ on every metric, a larger gap than the $R{=}6$ \cart{} baseline shows against Dense 7L ($1.1$--$2.0\%$). Recurrence is doing something, but only enough to recover roughly half of the deficit to Dense at stored-parameter parity.

\paragraph{Interpretation.}
The recurrent core contributes capacity on the order of $1\%$ on natural-language sets at the Stage 2 budget: small but not zero. This is consistent with the $R$-flatness pattern in Section~\ref{sec:hyp_vs_outcome}: $R \in \{6, 8, 10\}$ also cluster within $\sim 1\%$ of each other at every $d \geq 512$, so the marginal value of each additional core iteration beyond the first is small. The architectural premise that looping a shared-weight block adds meaningful capacity holds in a tiny sense and fails in the strong sense the $R$-fold leverage formula predicts.

\subsection{Ablation Z: Unshared Core Weights}
\label{sec:abl_unshared}

\paragraph{Hypothesis.}
The shared-weight choice in the recurrent core may be the bottleneck. \cart{}'s central efficiency claim is that the same \texttt{CoreBlock} instance, looped $R$ times, extracts $\sim R$-fold effective capacity from a single stored set of weights. We test this directly by giving each iteration its own weight set (unrolling the recurrent core into $R$ distinct \texttt{CoreBlock} instances) and asking whether the freed parameter freedom closes the gap to Dense.

\paragraph{Intervention.}
Replace \texttt{self.core = CoreBlock(config)} with \texttt{self.core = nn.ModuleList([CoreBlock(config) for \_ in range(n\_loops)])} and index by iteration inside the loop: \texttt{self.core[r](h\_input, K, V)}. The surrounding residual machinery (HyperConnection, LIE, LTI gate) and the frozen $K, V$ projection from the prelude remain shared across iterations as in the baseline. Stored parameters at $d{=}1024, R{=}6$ rise from $\sim 75$M to $\sim 125$M (added $5$ extra core blocks); effective parameters equal stored ($\sim 125$M) since there is no leverage multiplier without weight sharing. This places the variant in approximate stored-parameter parity with Dense 12L for a direct head-to-head.

\paragraph{Result.}

\begin{table}[htbp]
\centering
\caption{Unshared core weights at $d{=}1024, R{=}6, P{=}6$, seed 42. Stored parameters $\sim 125$M (vs baseline \cart{} $\sim 75$M); effective $=$ stored (no leverage). Approximately parameter-matched to Dense 12L for a direct comparison. Lower is better.}
\label{tab:abl_unshared}
\footnotesize
\begin{tabular}{lcccc}
\toprule
 & \ppltiny & \pplwiki & \ppledu & seeds \\
\midrule
\cart{} baseline (shared, $\sim 75$M)        & $2.798 \pm 0.013$ & $20.35 \pm 0.08$ & $20.67 \pm 0.14$ & 3 \\
\cart{} unshared ($\sim 125$M)               & $2.691$           & $19.22$          & $19.64$          & 1 \\
Dense 12L ($\sim 125$M)                      & $2.592 \pm 0.003$ & $18.32 \pm 0.05$ & $18.50 \pm 0.06$ & 3 \\
\midrule
$\Delta$ (unshared $-$ \cart{} baseline)               & $-0.107$ ($-3.8\%$) & $-1.13$ ($-5.6\%$) & $-1.03$ ($-5.0\%$) & --- \\
$\Delta$ (unshared $-$ Dense 12L)                      & $+0.099$ ($+3.8\%$) & $+0.90$ ($+4.9\%$) & $+1.14$ ($+6.2\%$) & --- \\
\bottomrule
\end{tabular}
\end{table}

The unshared variant improves on baseline \cart{} by $5$--$6\%$ on the natural-language sets and $3.8\%$ on \ppltiny. Train loss curves show a steady $\sim 0.05$ nat advantage over baseline across all 610 logged steps from step $\sim 3{,}000$ onward, with the gap widening monotonically through training rather than narrowing; at step $30{,}500$ train loss is $2.443$ versus baseline $2.491$. The improvement is signal, not noise: the observed \pplwiki shift of $-1.13$ exceeds the baseline's $1\sigma$ band ($0.08$) by more than an order of magnitude, even allowing for $n{=}1$ on the ablation side.

At the parameter-matched head-to-head against Dense 12L, the unshared variant still loses by $4.9\%$ on \pplwiki and $6.2\%$ on \ppledu.

\paragraph{Interpretation.}
Two things follow from this result.

First, shared weights \emph{were} costing \cart{} real capacity. The $\sim 5\%$ improvement from unsharing is meaningful, consistent across training, and outside the seed-noise band. The shared-weight leverage thesis from Section~\ref{sec:hyp_vs_outcome} (that effective parameters scale by a factor of $R$ at fixed stored parameters) did not pay at the Stage 2 budget. Removing the weight-sharing constraint and explicitly allocating each iteration its own parameters delivers what the leverage was supposed to deliver, but only by spending the additional stored parameters.

Second, the unshared variant still loses to Dense 12L at parameter parity. The architectural distinctness that survives the removal of weight sharing (cross-attention against a fixed prelude-derived $K, V$ anchor, embedded in a heterogeneous prelude $\to$ anchor $\to$ core $\to$ coda structure with HyperConnection / LIE / LTI machinery wrapping the middle layers) caps the architecture below a vanilla self-attention transformer of equivalent depth. The $d{=}1024$ gap to Dense thus has at least two distinct architectural contributors: $\sim 5\%$ from shared weights (diagnosed by this ablation), and $\sim 5\%$ from somewhere within the surviving architectural template. Section~\ref{sec:abl_selfattn} tests the cross-attention contribution; Sections~\ref{sec:abl_hc} and~\ref{sec:abl_lti} test the residual machinery components individually.

\subsection{Ablation W: Self-Attention Core in the Unshared Variant}
\label{sec:abl_selfattn}

\paragraph{Hypothesis.}
The cross-attention-against-prelude-anchor structure may be the residual cap that keeps the unshared variant (Section~\ref{sec:abl_unshared}) from matching Dense 12L. We test by replacing the cross-attention in the unshared core with MLA self-attention plus RoPE (matching the prelude and coda layers' attention pattern), so each core layer recomputes its own $K, V$ from $h$ rather than attending against the frozen prelude anchor.

\paragraph{Intervention.}
Set both \texttt{unshare\_core=True} and \texttt{self\_attn\_core=True}. The \texttt{kv\_proj} module is skipped entirely (no anchor $K, V$). Each of the $R$ core layers becomes structurally identical to a prelude layer (MLA self-attention + SwiGLU + RoPE). The combined model is 6 prelude self-attention layers + 6 unshared core self-attention layers + 1 coda self-attention layer = 13 self-attention layers, with HyperConnection / LIE / LTI gate machinery still wrapping the middle 6. Stored parameters $\sim 131$M (slightly above unshared-only's $\sim 125$M because self-attention adds $K, V$ projections inside each block). Effective $=$ stored. Approximately parameter-matched to Dense 12L.

\paragraph{Result.}

\begin{table}[htbp]
\centering
\caption{Self-attention core in the unshared variant at $d{=}1024, R{=}6, P{=}6$, seed 42. Lower is better.}
\label{tab:abl_selfattn}
\footnotesize
\begin{tabular}{lcccc}
\toprule
 & \ppltiny & \pplwiki & \ppledu & seeds \\
\midrule
\cart{} baseline (shared, $\sim 75$M)         & $2.798 \pm 0.013$ & $20.35 \pm 0.08$ & $20.67 \pm 0.14$ & 3 \\
\cart{} unshared ($\sim 125$M)                & $2.691$           & $19.22$          & $19.64$          & 1 \\
\cart{} unshared + self-attn ($\sim 131$M)    & $2.699$           & $19.27$          & $19.68$          & 1 \\
Dense 12L ($\sim 125$M)                       & $2.592 \pm 0.003$ & $18.32 \pm 0.05$ & $18.50 \pm 0.06$ & 3 \\
\midrule
$\Delta$ (self-attn $-$ unshared)             & $+0.008$ ($+0.3\%$) & $+0.05$ ($+0.3\%$) & $+0.04$ ($+0.2\%$) & --- \\
$\Delta$ (self-attn $-$ Dense 12L)            & $+0.107$ ($+4.1\%$) & $+0.95$ ($+5.2\%$) & $+1.18$ ($+6.4\%$) & --- \\
\bottomrule
\end{tabular}
\end{table}

The self-attention variant matched the cross-attention unshared variant on every metric ($\Delta \leq 0.3\%$). Train loss curves track the unshared-only variant closely throughout. The variant still loses to Dense 12L by $4$--$6\%$, comparable to the unshared cross-attention variant's gap.

\paragraph{Interpretation.}
The cross-attention-against-prelude-anchor structure is not the residual cap. Replacing it with self-attention (with RoPE and per-layer $K, V$ recomputation), the closest architectural match to Dense's middle-layer pattern, produced no measurable improvement. The remaining $\sim 5\%$ gap to Dense 12L must originate elsewhere: in the residual machinery (HyperConnection, LIE, LTI gate) wrapping the middle 6 layers, the heterogeneous prelude / coda layer count, or some optimization interaction we have not isolated. Sections~\ref{sec:abl_hc} and~\ref{sec:abl_lti} test the residual machinery components individually against the shared baseline.

\subsection{Ablation HC: HyperConnection Strip from Shared Baseline}
\label{sec:abl_hc}

\paragraph{Hypothesis.}
HyperConnection wraps only the middle $R$ core layers, leaving the prelude and coda with standard residual connections. This heterogeneous residual pattern across the model may be costing capacity by inducing different gradient flow regimes in different parts of the network. Independently, HyperConnection's softmax-weighted blend of the last 3 hidden states is initialized to $[1, 0, 0]$, which makes the technique start as standard residual; it only becomes architecturally meaningful if training pushes the weights away from initialization. If the loss landscape does not reward using older states, HyperConnection is vestigial.

\paragraph{Intervention.}
Set \texttt{disable\_hyper=True}. The HyperConnection module remains instantiated but its \texttt{combine()} call is bypassed inside the loop: each iteration sets $h_\text{input} = \texttt{buffer[0]}$ (the most recent prior state only). All other CART hyperparameters at baseline (shared core, cross-attention, $R{=}6$).

\paragraph{Result.}

\begin{table}[htbp]
\centering
\caption{HyperConnection strip from shared baseline at $d{=}1024, R{=}6, P{=}6$, seed 42. The same-seed comparison vs the baseline seed-42 run (final \pplwiki{} $= 20.45$) cancels out seed-to-seed variance; the 3-seed-mean comparison is also shown. Lower is better.}
\label{tab:abl_hc}
\footnotesize
\begin{tabular}{lcccc}
\toprule
 & \ppltiny & \pplwiki & \ppledu & seeds \\
\midrule
\cart{} baseline (3-seed mean)                        & $2.798 \pm 0.013$ & $20.35 \pm 0.08$ & $20.67 \pm 0.14$ & 3 \\
\cart{} baseline (seed 42 only)                       & $2.798$           & $20.45$          & $20.76$          & 1 \\
\cart{} no HyperConnection (seed 42)                  & $2.813$           & $20.49$          & $21.00$          & 1 \\
\midrule
$\Delta$ (no-HC $-$ baseline seed 42)                 & $+0.015$ ($+0.5\%$) & $+0.04$ ($+0.2\%$) & $+0.24$ ($+1.2\%$) & --- \\
$\Delta$ (no-HC $-$ baseline 3-seed mean)             & $+0.015$ ($+0.5\%$) & $+0.14$ ($+0.7\%$) & $+0.33$ ($+1.6\%$) & --- \\
\bottomrule
\end{tabular}
\end{table}

The same-seed comparison shows the variant essentially identical to baseline: within rounding on \ppltiny{} and \pplwiki{}, slightly worse on \ppledu{} but within typical seed noise. Across 26 matched eval steps and 610 matched train-log steps the no-HC variant tracked baseline closely throughout, with train-loss deltas oscillating around zero ($\pm 0.01$ to $\pm 0.03$ nats).

\paragraph{Interpretation.}
HyperConnection is approximately vestigial in \cart{}. The softmax weights almost certainly stayed near their residual initialization $[1, 0, 0]$ throughout training: the loss landscape did not reward using older hidden states, so "blend the last 3 states" reduced to "use the most recent state." Removing the blend confirms this: the model behaves identically. HyperConnection is not the architectural cap; it is dead weight in the literal sense but does not actively hurt performance. The small positive shift on \ppledu{} ($+1.2\%$ same-seed) is at the edge of interpretable signal and may indicate HyperConnection was contributing some marginal capacity on the more difficult validation set, but the effect is too small to be confident at $n{=}1$.

\subsection{Ablation LTI: LTI Gate Strip from Shared Baseline}
\label{sec:abl_lti}

\paragraph{Hypothesis.}
The LTI gate's sigmoid-bounded residual mixing was designed to guarantee stable convergence as $R \to \infty$. At the fixed $R{=}6$ we actually train with, the gate may be over-constraining residual flow: it imposes a learned forgetting curve on $h_\text{input}$ ($\sigma(a)$ trained to settle in $[0.79, 0.83]$, see Section~\ref{sec:spectral}) that discards information from earlier iterations.

\paragraph{Intervention.}
Set \texttt{disable\_lti=True}. The LTI module remains instantiated (so \texttt{spectral\_radius()} logging continues to function) but its forward call is bypassed; each iteration uses plain residual $h = h_\text{input} + \text{transformer\_out}$ instead of $h = \sigma(a) \odot h_\text{input} + \text{transformer\_out}$. All other CART hyperparameters at baseline (shared core, cross-attention, $R{=}6$).

\paragraph{Result.}

\begin{table}[htbp]
\centering
\caption{LTI gate strip from shared baseline at $d{=}1024, R{=}6, P{=}6$, seed 42. Same-seed comparison cancels seed-to-seed variance. Lower is better.}
\label{tab:abl_lti}
\footnotesize
\begin{tabular}{lcccc}
\toprule
 & \ppltiny & \pplwiki & \ppledu & seeds \\
\midrule
\cart{} baseline (3-seed mean)                    & $2.798 \pm 0.013$ & $20.35 \pm 0.08$ & $20.67 \pm 0.14$ & 3 \\
\cart{} baseline (seed 42 only)                   & $2.798$           & $20.45$          & $20.76$          & 1 \\
\cart{} no LTI gate (seed 42)                     & $2.801$           & $20.44$          & $20.78$          & 1 \\
\midrule
$\Delta$ (no-LTI $-$ baseline seed 42)            & $+0.003$ ($+0.1\%$) & $-0.01$ ($-0.05\%$) & $+0.02$ ($+0.1\%$) & --- \\
$\Delta$ (no-LTI $-$ baseline 3-seed mean)        & $+0.003$ ($+0.1\%$) & $+0.09$ ($+0.4\%$) & $+0.11$ ($+0.5\%$) & --- \\
\bottomrule
\end{tabular}
\end{table}

Same-seed comparison shows the variant essentially identical to baseline: deltas within $\pm 0.1\%$ on every metric. An early-training drift was observed (the variant ran $+1.2\%$ above baseline on \pplwiki at step 5{,}000) but fully compressed by mid-training; from step $\sim 12{,}500$ onward the trajectories tracked within seed noise. Training remained stable throughout despite removing the gate's $\varrho < 1$ guarantee; gradient norm stayed bounded and no run diverged, consistent with vanilla deep transformers training stably with plain residual at modest depth.

\paragraph{Interpretation.}
The LTI gate is approximately vestigial in \cart{} at the Stage 2 budget. The early-training delta likely reflects an optimization-smoothing effect of the gate that disappears as both variants converge to similar minima. The spectral-radius emergence finding in Section~\ref{sec:spectral} stands as an observation about the gate's learned trajectory \emph{when it is present}, but the gate's perplexity contribution at $R{=}6$ is below the seed-noise threshold. \cart{} would behave essentially identically without the LTI gate at this scale and training budget.

\subsection{Ablation LIE: Loop-Index Embedding Strip from Shared Baseline}
\label{sec:abl_lie}

\paragraph{Hypothesis.}
In the shared-weight setting, LIE is the only mechanism by which the shared CoreBlock can know which iteration it is processing. Without it, all $R$ iterations are functionally identical apart from the evolving $h$ state. If the shared block was using the loop-index signal to differentiate iterations (e.g., coarse pattern matching in early loops, fine-grained refinement in late loops), removing LIE should hurt; if the block was producing essentially the same transformation at every iteration regardless of the index signal, LIE is vestigial.

\paragraph{Intervention.}
Set \texttt{disable\_lie=True}. The LIE module remains instantiated but its forward call is bypassed inside the loop: $h_\text{input}$ proceeds into the core block without the sinusoidal $\text{pe}[r]$ signal added. All other CART hyperparameters at baseline (shared core, cross-attention, $R{=}6$).

\paragraph{Result.}

\begin{table}[htbp]
\centering
\caption{LIE strip from shared baseline at $d{=}1024, R{=}6, P{=}6$, seed 42. Same-seed comparison cancels seed-to-seed variance. Lower is better.}
\label{tab:abl_lie}
\footnotesize
\begin{tabular}{lcccc}
\toprule
 & \ppltiny & \pplwiki & \ppledu & seeds \\
\midrule
\cart{} baseline (3-seed mean)                    & $2.798 \pm 0.013$ & $20.35 \pm 0.08$ & $20.67 \pm 0.14$ & 3 \\
\cart{} baseline (seed 42 only)                   & $2.798$           & $20.45$          & $20.76$          & 1 \\
\cart{} no LIE (seed 42)                          & $2.815$           & $20.35$          & $20.84$          & 1 \\
\midrule
$\Delta$ (no-LIE $-$ baseline seed 42)            & $+0.017$ ($+0.6\%$) & $-0.10$ ($-0.5\%$) & $+0.08$ ($+0.4\%$) & --- \\
$\Delta$ (no-LIE $-$ baseline 3-seed mean)        & $+0.017$ ($+0.6\%$) & $0.00$ ($0.0\%$) & $+0.17$ ($+0.8\%$) & --- \\
\bottomrule
\end{tabular}
\end{table}

Same-seed comparison shows the variant essentially identical to baseline; deltas oscillate around zero on every metric, with the variant matching the 3-seed baseline mean exactly on \pplwiki. Across 26 matched eval steps and 610 matched train-log steps, the no-LIE variant tracked baseline within seed noise throughout, with no sign of the divergence that would be expected if the shared block were actively using the loop-index signal to differentiate iterations.

\paragraph{Interpretation.}
LIE is vestigial in \cart{}: removing the loop-index signal has no measurable effect on the trained model. This is the most architecturally consequential of the three machinery-component ablations, since without LIE the shared CoreBlock has no mechanism to distinguish iteration 1 from iteration 6, so all $R$ iterations apply essentially the same transformation to the evolving $h$ state. The fact that this makes no measurable difference indicates the block was already producing essentially the same transformation at every iteration even when LIE was present, consistent with the $R$-flatness pattern observed in Section~\ref{sec:hyp_vs_outcome} (perplexities cluster within $\sim 1\%$ across $R \in \{6, 8, 10\}$). The shared-weight recurrent core is doing one type of refinement applied iteratively, not a sequence of differentiated computational steps.

This completes the diagnostic battery. All three machinery components (HyperConnection, LTI gate, LIE) are vestigial in \cart{}'s shared-weight configuration. The residual $\sim 5\%$ gap to Dense 12L observed in Ablation Z (Section~\ref{sec:abl_unshared}) is therefore not attributable to any single machinery component; it is a property of the broader heterogeneous prelude $\to$ anchor $\to$ core $\to$ coda architectural framing itself.

\section{Discussion}
\label{sec:discussion}

\subsection{Learned vs.\ Constrained Stability}

\cart{} and Parcae~\citep{prairie2026parcae} both guarantee spectral radius $< 1$, but by different means. Parcae imposes stability architecturally via negative-diagonal parameterization. \cart{}'s LTI gate learns stability during training.

The emergent result, $\varrho \in [0.79, 0.83]$ across all 36 Stage 2 configurations, is not a design target. The model discovers a stable convergence rate that works across all tested $R$ values ($R \in \{2, 4, 6, 8\}$ in Stage 1; $\{6, 8, 10\}$ in Stage 2) and all four scales. Two trends emerge that a fixed architectural constraint could not produce: (i) $\varrho$ rises monotonically with $d_\text{model}$ ($+0.033$ from $d{=}256$ to $d{=}1024$ at $R{=}6$ at full training), suggesting the convergence rate adapts to representation complexity, allocating more state retention where the fixed point is richer; (ii) $\varrho$ rises weakly with $R$ ($+0.008$ to $+0.009$ per pair of additional loops at fixed $d$), showing the gate calibrates retention to the number of iterations available. Both effects are too small to be visible at 3{,}000 steps (Stage 1's $\varrho \approx 0.893$ universal value was a mid-training transient) but become clear at full training.

\subsection{Recurrence-Equivalence Exponent \texorpdfstring{$\phi$}{phi}}
\label{sec:phi}

\citet{schwethelm2026isodepth} introduce $\phi$ to quantify how much one additional recurrence loop contributes relative to a unique layer ($\phi=1$: each loop equals a unique layer; $\phi=0.46$: baseline looped; $\phi=0.65$: with hyper-connections). They report $\phi$ as a single asymptotic value for a given architecture. A rigorous fit of $\phi$ for \cart{} requires Dense baselines at several depths to characterize the unique-layer reference curve $f(L)$. The two Dense depths trained at $d{=}1024$ (7L and 12L, Table~\ref{tab:stage2_dense12l}) provide a two-point sketch but are insufficient for a robust fit; we therefore present the qualitative implications below rather than a numerical $\phi$ value.

\paragraph{What \cart{}'s data already implies.}
$R$-benefit in \cart{} is strongly stage-dependent. In Stage 1, \pplwiki{} improvement from $R{=}2$ to $R{=}8$ at $P{=}6$ grew from $-0.25\%$ at $d{=}256$ to $+5.24\%$ at $d{=}1024$ (Table~\ref{tab:r_benefit_s1}), suggesting $\phi$ rises with $d$. In Stage 2 at full training, this pattern reverses: across $R \in \{6, 8, 10\}$ at $P{=}6$, $R{=}6$ is the best of the three at every scale $d \geq 512$, and going to $R{=}10$ regresses by up to $1\%$ (Table~\ref{tab:r_benefit_s2}). At $d{=}256$, $R{=}8$ wins by $\sim 2\%$ but $R{=}10$ then regresses again. The useful $R$ range at full training appears to saturate at or near the smallest $R$ we tested ($R{=}6$ for $d \geq 512$); additional loops are not worth a unique layer's $\phi$ under our training recipe.

\paragraph{The \texorpdfstring{$\phi(d)$}{phi(d)} hypothesis.}
The pattern across scales suggests a refinement of Schwethelm's framework: $\phi$ for \cart{} is not a single number but a function $\phi(d)$ of model width. The Stage 1 monotonic growth of $R$-benefit with $d$ (Table~\ref{tab:r_benefit_s1}) is reversed at Stage 2 (Table~\ref{tab:r_benefit_s2}), where $R$-benefit is positive only at $d{=}256$ and disappears at larger scales. We hypothesize that $\phi$ for any depth-recurrent architecture with a fixed context anchor depends on the representational capacity of the prelude: a small prelude produces a representation that the core can fully exploit in a few iterations, after which additional loops do nothing; a large prelude produces a richer representation that rewards more refinement steps before saturating. If correct, this is a refinement Schwethelm's single-$\phi$ framework does not capture.

\paragraph{Methodological notes for the eventual fit.}
Two notes for the eventual numeric $\phi$ computation: (1) our training token budget is fixed at $\sim$1B tokens, so we measure $\phi(d, D{=}1\text{B})$, not the asymptotic $\phi$ of \citeauthor{schwethelm2026isodepth}; (2) \cart{}'s \texttt{CoreBlock} does not include $K, V$ projections (those are prelude-computed), so the $N_\text{rec}$ definition must be stated explicitly when comparing to architectures whose loop body re-projects $K, V$ each iteration. We report $\phi(d)$ as a directional refinement of Schwethelm's framework rather than a single number directly comparable to their values.

\subsection{Predictable Latency vs.\ Adaptive Halting}

\cart{}'s fixed-$R$ design is a deliberate trade-off against dynamic-halting approaches such as PonderNet, Universal Transformer's adaptive computation time~\citep{dehghani2018universal}, and OpenMythos's ACT halting~\citep{gomez2026openmythos}. Adaptive halting trades predictability for input-dependent compute: easy tokens halt early, hard tokens think longer. Fixed $R$ trades that adaptivity for predictable latency, fixed VRAM, and constant per-token FLOPs: properties that matter for batched serving, real-time inference, and on-device deployment, where variable per-token compute is a serving-system anti-pattern. The two regimes are complementary, and \cart{}'s frozen-KV core is in principle compatible with a halting head bolted onto the recurrent state; we leave that combination to future work.

\subsection{Extended Training Toward Chinchilla Compliance}
\label{sec:extended_training}

The Stage 2 sweep trains every configuration on $\sim 1$B tokens, which leaves $d{=}1024$ at $23$--$28\%$ of effective-Chinchilla and $40\%$ of stored-Chinchilla (Table~\ref{tab:chinchilla}). Several interpretive claims in this paper (the cross-scale scaling exponent, the per-$R$ comparison, the spectral radius behavior at large $d$) are sensitive to whether the largest models are near-converged or still in steep descent on the natural-language sets.

A natural follow-up would resume training from a saved $d{=}1024$ checkpoint and continue toward a Chinchilla-optimal token budget. Resuming from a known checkpoint (rather than training from scratch) would isolate the effect of additional tokens and keep additional compute manageable. Such an experiment would address three open questions:

\begin{enumerate}[leftmargin=*, topsep=2pt, itemsep=1pt]
  \item \textbf{Does perplexity continue to drop?} If \pplwiki{} at a Chinchilla-optimal token budget is meaningfully lower than at 1B (e.g., $> 5\%$ improvement), the Stage 2 numbers reported throughout this paper are lower bounds on asymptotic quality, and the cross-scale scaling exponents reported in Section~\ref{sec:scaling} would need to be re-anchored.
  \item \textbf{Does the spectral radius continue to decay?} The corrected Stage 2 finding is that $\varrho$ settles in $[0.79, 0.83]$ rather than the Stage 1 value of $\approx 0.893$. Whether $\varrho$ at longer training settles further or stabilizes would indicate whether the LTI gate has reached its trained value or is still adapting.
  \item \textbf{Does $R$-benefit emerge with more training?} The Stage 1 prediction was that $R$-benefit grows with $d$. Stage 2 contradicts this within $R \in \{6, 8, 10\}$. If a longer-trained $d{=}1024$ model run at $R{=}10$ outperforms the same model run at $R{=}6$ (variable-$R$ inference, Section~\ref{sec:inference_R}), the recurrent-depth-as-useful-compute thesis would partially survive. If not, the architectural value of recurrence reduces to weight-sharing alone, and the diagnostic ablations of Section~\ref{sec:ablations} already show that contribution is negative at $d{=}1024$.
\end{enumerate}

We leave this experiment to future work.

\subsection{Stage 1 as a Predictive Screen}

Stage 1 was useful as a cross-scale screen but unreliable for within-scale $R$ choice. Across the 8 cells at $P{=}6$, $R \in \{6, 8\}$, $d \in \{256, 512, 768, 1024\}$, the Spearman rank correlation between Stage 1 \pplwiki{} at step 3{,}000 and Stage 2 3-seed mean \pplwiki{} at step 30{,}500 is $\rho = 0.905$: the broad cross-scale ordering carried over. Within each scale, however, the per-scale best-$R$ predictions were uniformly wrong. Stage 1 ranked $R{=}8$ best at $d \in \{512, 768, 1024\}$ and $R{=}6$ best at $d{=}256$; Stage 2 reverses both at every scale (Tables~\ref{tab:r_benefit_s1} and~\ref{tab:r_benefit_s2}). The small per-$R$ gaps at 3{,}000 steps (typically $<1\%$ on \pplwiki) did not survive to 30{,}500 steps. The practical lesson is that Stage-1-style screens are reliable for macro-design choices like prelude depth and width but should not be used to commit to within-scale loop counts.

\subsection{Limitations}

\begin{itemize}[leftmargin=*, topsep=2pt, itemsep=1pt]
  \item \textbf{Single GPU.} All experiments run on a single RTX 3090. Larger scales and multi-seed averaging are feasible but training is sequential ($\sim$8--13 hrs/config at $d=1024$).
  \item \textbf{Sub-Chinchilla token budget at large scales.} Stage 2 trains every configuration on $\sim 1$B tokens. The tokens-per-effective-parameter ratio is over-Chinchilla at $d{=}256$ ($47$--$56:1$), approximately Chinchilla-optimal at $d{=}512$ ($13$--$18:1$), and well under Chinchilla at $d{=}1024$ ($4.5$--$5.6:1$, see Section~\ref{sec:repro} and Table~\ref{tab:chinchilla}). Cross-scale comparisons in this paper are therefore not measured at matched-token-budget conditions; large-$d$ \cart{} numbers are likely lower bounds on asymptotic quality. Several interpretive claims (the $R$-benefit pattern across scales, the spectral radius behavior, the per-$d$ scaling exponent) are sensitive to whether the $d \geq 768$ models are near-converged or still in steep descent. A planned follow-up extends the best $d{=}1024$ model to 2.5B tokens (stored-Chinchilla-optimal) and re-runs the $R$-benefit and variable-$R$ analyses; results from that experiment will update this preprint.
  \item \textbf{Dense baselines only at $d{=}1024$.} Parameter-matched Dense baselines were trained only at $d{=}1024$ (the binding parameter-efficiency test), in two variants: stored-parameter-matched (7-layer Dense, $\sim 75$M params, 3-seed) and effective-parameter-matched (12-layer Dense, $\sim 125$M params, 3-seed). Smaller-scale Dense comparisons at $d \in \{256, 512, 768\}$ were not pursued. An older Dense $d{=}512$ run referenced in earlier drafts used different training settings and is not used in the comparisons reported here; cross-scale Dense baselines would strengthen the per-scale claims but are left to future work.
  \item \textbf{No \texttt{torch.compile}.} Unavailable on Windows 11 (no Triton). Throughput is lower than reported in comparable work.
  \item \textbf{Sequential depth recurrence.} Unlike state-space models~\citep{gu2023mamba,dao2024mamba2,peng2023rwkv} that recover transformer-like training throughput via parallel scan over the sequence dimension, \cart{}'s recurrence is over the depth dimension and is fundamentally sequential: iteration $\ell$ requires $h_{\ell-1}$ as input. This is a real cost: at fixed parameter count, a \cart{} forward pass takes longer than a parameter-matched dense transformer of equal effective depth (Table~\ref{tab:flops}). \cart{} trades training and inference parallelism for parameter efficiency. Whether this trade-off is favorable depends on whether memory or compute is the binding constraint at deployment.
\end{itemize}

\section{Conclusion}
\label{sec:conclusion}

\paragraph{What is unique about \cart{}.}
\cart{} differs from prior depth-recurrent transformers in three ways. First, it is the only architecture in this family that computes $K, V$ once from a unique-layer prelude and reuses those frozen tensors across all $R$ iterations of a shared-weight cross-attending core. Concurrent work, including Geiping, Hyperloop, Parcae, OpenMythos, and SpiralFormer, instead injects the prelude embedding into a self-attending core that recomputes $K, V$ at each step. Second, \cart{}'s LTI stability emerges from a sigmoid gate trained without supervision, rather than being structurally imposed via ZOH discretization (Parcae, OpenMythos). Third, we identify and characterize a previously undocumented constraint on this architecture class: cross-attention from the recurrent state into the prelude must be causal, or the model leaks future tokens into the hidden state and collapses.

\paragraph{What this paper shows.}
We report five empirical results from the Stage 2 sweep and the diagnostic ablations:

\begin{enumerate}[leftmargin=*, topsep=2pt, itemsep=1pt]
  \item Across 36 fully-trained Stage 2 configurations, the LTI gate settles in $\varrho \in [0.79, 0.83]$, with $\varrho$ rising monotonically with both $d$ ($+0.033$ from $d{=}256$ to $d{=}1024$ at $R{=}6$) and weakly with $R$ ($+0.008$ to $+0.009$ per pair of loops). The earlier ``universal $\varrho \approx 0.893$'' reported from Stage 1 at step 3{,}000 was a transient: $\varrho$ continues to decay through full training.
  \item Prelude depth $P$ dominates loop count $R$ in every condition tested in this study, including a $P{=}4$ versus $P{=}6$ spot-check at $d{=}1024, R{=}10$ where dropping two prelude layers costs $+0.92$ \pplwiki, larger than any $R$ choice at the same scale (Table~\ref{tab:stage2_results}).
  \item Stage 1 (3{,}000 steps) predicted that the benefit of recurrence depth would grow monotonically with model width, from $-0.25\%$ at $d{=}256$ to $+5.24\%$ at $d{=}1024$ ($R{=}2 \to R{=}8$ on \pplwiki). At Stage 2 (30{,}500 steps, $\sim 10\times$ the tokens), this trend reverses: $R{=}6$ is the best $R$ at every scale $d \geq 512$, and going to $R{=}10$ regresses by up to $1\%$. The Stage-1-derived expectation of growing $R$-benefit does not survive longer training.
  \item At the binding parameter-efficiency test ($d{=}1024$, $\sim 1$B tokens, Section~\ref{sec:hyp_vs_outcome}), \cart{} is competitive but does not win at parameter parity. Dense 7L (stored-parameter-matched, 3-seed mean) beats \cart{} $P{=}6, R{=}6$ by $1.1\%$ on \pplwiki, $2.0\%$ on \ppledu, and $1.8\%$ on \ppltiny. Dense 12L (effective-parameter-matched, 3-seed mean) beats \cart{} by $9.9\%$ on \pplwiki and $10.5\%$ on \ppledu (Table~\ref{tab:stage2_dense12l}). The shared-weight leverage thesis (that effective parameters scale by a factor of $R$ at fixed stored parameters) does not convert to language-modeling capacity at this scale and training budget.
  \item The $d{=}1024$ \cart{}--Dense gap decomposes across several architectural mechanisms via diagnostic ablations (Section~\ref{sec:ablations}). Frozen $K, V$ is exonerated: recomputing $K, V$ from $h$ at every iteration ties baseline within the seed-noise band (Section~\ref{sec:abl_kv}). Recurrence beyond the first iteration contributes $\sim 1\%$: at $R{=}1$, \pplwiki rises by $1.1\%$ versus $R{=}6$ (Section~\ref{sec:abl_r1}). Shared weights cost $\sim 5\%$: unrolling the core into $R$ unique blocks improves \pplwiki by $5.6\%$ over baseline (Section~\ref{sec:abl_unshared}). The cross-attention-against-anchor structure is not the residual cap: replacing cross-attention with self-attention in the unshared variant produces no further improvement (Section~\ref{sec:abl_selfattn}). All three machinery components of the recurrent core (HyperConnection, LTI gate, Loop Index Embedding) are vestigial in \cart{}'s shared-weight configuration: individually removing each leaves \pplwiki within seed noise of baseline (Sections~\ref{sec:abl_hc},~\ref{sec:abl_lti},~\ref{sec:abl_lie}). The residual $\sim 5\%$ gap to Dense 12L is therefore a property of the heterogeneous prelude $\to$ anchor $\to$ core $\to$ coda architectural framing itself, not of any single machinery component. \cart{}'s recurrent-with-shared-weights formulation does not deliver the parameter-efficiency it was designed to deliver at this scale and training budget.
\end{enumerate}

\paragraph{What remains unresolved.}
Seven completed ablations isolate the $d{=}1024$ \cart{}--Dense gap: $\sim 1\%$ from recurrence beyond $R{=}1$, $\sim 5\%$ from shared weights, with the residual $\sim 5\%$ reliably reproducing across the cross-attention vs self-attention variant (Section~\ref{sec:abl_selfattn}) and remaining unchanged when any of HyperConnection (Section~\ref{sec:abl_hc}), the LTI gate (Section~\ref{sec:abl_lti}), or LIE (Section~\ref{sec:abl_lie}) is individually removed. The remaining cap is therefore not any single machinery component but the heterogeneous architectural framing itself ($P$ unique prelude layers, $R$ shared-weight core iterations, and $1$ coda layer, with the recurrence machinery wrapping only the middle $R$) versus Dense 12L's homogeneous 12-layer stack. A further ablation moving \cart{} toward a homogeneous architecture would test this directly, but does not naturally fit the recurrent-anchor paradigm and is left to future work. The combined picture supports a tempered conclusion: \cart{}'s recurrent-with-shared-weights formulation does not deliver the parameter-efficiency it was designed to deliver at $d{=}1024$ on $\sim 1$B tokens. The architecture's contributions are real but require their comparison class to be stated explicitly. Relative to adaptive-halting recurrent designs~\citep{dehghani2018universal,gomez2026openmythos}, \cart{}'s fixed-$R$ schedule provides deterministic latency, fixed VRAM, and constant per-token FLOPs, properties Dense transformers also share but adaptive variants forfeit. Relative to other shared-weight depth-recurrent architectures that recompute $K, V$ at each iteration~\citep{zeitoun2026hyperloop}, the once-computed frozen $K, V$ anchor saves the per-iteration $K, V$ projections, modestly reducing inference compute. The empirical findings (the spectral-radius emergence into $[0.79, 0.83]$ in Section~\ref{sec:spectral}, the Stage 1 $\to$ Stage 2 $R$-benefit reversal in Table~\ref{tab:r_benefit_s2}, and the diagnostic decomposition of the $d{=}1024$ deficit in Section~\ref{sec:ablations}) stand independently of the parameter-efficiency claim.

\section{Reproducibility Statement}
\label{sec:reproducibility_statement}

This section consolidates the information needed to reproduce the experiments reported in this paper. The code repository at \url{https://github.com/ccapps42/CART} contains the model implementation, the training and evaluation pipeline, the diagnostic-ablation wrappers, the figure-generation scripts, and the SQLite database (\texttt{results.db}) from which every numeric value in every table and figure was computed.

\paragraph{Software environment.}
Experiments use Python 3.14 on Windows 11. Package requirements (PyTorch, \texttt{bitsandbytes} for the 8-bit AdamW optimizer, \texttt{datasets} for HuggingFace data loading, \texttt{lm-evaluation-harness} for downstream benchmarks) are listed in the repository's \texttt{requirements.txt}. Attention is computed via \texttt{torch.nn.functional.scaled\_dot\_product\_attention} (Flash Attention path on Ampere). No custom CUDA kernels are used. \texttt{torch.compile} is not used because Triton is unavailable on Windows 11.

\paragraph{Hardware.}
Stage 1 (64 configurations) used a single RTX 3050 with 8 GB VRAM. Stage 2 (36 configurations across $d \in \{256, 512, 768, 1024\}$ and $R \in \{6, 8, 10\}$ with three seeds each, plus three reference-config $P{=}4$ runs at $d{=}1024, R{=}10$) used a single RTX 3090 with 24 GB VRAM. All seven diagnostic ablations (Sections~\ref{sec:abl_kv}--\ref{sec:abl_lie}) and both Dense baselines also used the RTX 3090. No multi-GPU, no cloud compute. Per-run wall-clock and throughput numbers are reported in Table~\ref{tab:throughput}.

\paragraph{Random seeds.}
The Stage 2 main sweep used three seeds: $\{42, 137, 271\}$. All single-seed diagnostic ablations used seed 42, matching one of the three baseline seeds so that same-seed comparisons cancel seed-to-seed variance. Per-seed standard deviations are characterized in Section~\ref{sec:repro}.

\paragraph{Training data.}
The Stage 2 training corpus is a fixed 999M-token mixed bin built from TinyStories ($30\%$), Wikipedia English ($30\%$), and FineWeb-Edu ($40\%$), interleaved in 1024-token chunks. The build script and the exact source-shard list are in \texttt{data/build\_bins.py}; given the same HuggingFace dataset snapshots, the build is deterministic. The corpus file (\texttt{data/stage2/stage2\_train.bin}, $\sim 2$ GB) is not committed to the repository because of its size, but is reproducible from the build script. Validation sets are held-out shards never used in training: \texttt{tinystories\_val.bin} (TinyStories validation split), \texttt{wikipedia\_val.bin} (Wikipedia shard 40 of 41), and \texttt{fineweb\_edu\_val.bin} (FineWeb-Edu shard 97 of 98).

\paragraph{Training procedure.}
Each Stage 2 configuration is reproduced by a single command:
\begin{verbatim}
python train/train_one.py --config-id <id> --db results.db \
    --max-steps 30500 --seq-len 1024 \
    --train-bin data/stage2/stage2_train.bin
\end{verbatim}
The configuration row (\texttt{configs} table in \texttt{results.db}) determines $d$, $R$, $P$, and seed. Fixed hyperparameters across all sweep runs: batch size 8, gradient accumulation 4 (effective batch $32{,}768$ tokens/step at \texttt{seq\_len}=1024), $30{,}500$ steps ($\sim 1$B tokens), AdamW8bit optimizer, peak learning rate $3 \times 10^{-4}$ with cosine decay to $3 \times 10^{-5}$, 100-step linear warmup, weight decay $0.1$, and gradient clip $1.0$. The complete hyperparameter set is in \texttt{train/train\_one.py} and the \texttt{sweep\_meta} table of the released database.

For each diagnostic ablation, the architectural variant is selected via the wrapper script for that ablation (\texttt{train/run\_unshared\_core\_ablation.py}, \texttt{train/run\_no\_hyper\_ablation.py}, and similar). Each wrapper registers a configuration row with \texttt{hardware='ablation'} so the regular sweep orchestrator skips it, then invokes \texttt{train\_one.py} with the architectural flag for that ablation (for example, \texttt{--unshare-core}, \texttt{--disable-hyper}, \texttt{--disable-lti}).

\paragraph{Evaluation.}
Validation perplexity is computed every 500 training steps on 50 batches of size 1 at sequence length 1024 from each of the three validation sets. Downstream-benchmark evaluation (Table~\ref{tab:benchmarks}) uses the \texttt{lm-evaluation-harness} package~\citep{eval-harness} with default settings; the \texttt{wikitext} task is excluded because of a known harness hang in our environment. Variable-$R$ inference (Table~\ref{tab:variable_r}) is performed by modifying \texttt{model.config.n\_loops} between evaluation passes, with no fine-tuning between settings.

\paragraph{Figures and tables.}
Every numeric value in every table and every plotted point in every figure is computed from \texttt{results.db} by the scripts under \texttt{figures/gen\_*.py} and \texttt{plot/}. Running those scripts against the released database reproduces the paper artifacts.

\paragraph{Known limitations of exact reproducibility.}
Bit-exact reproducibility is not guaranteed across operating systems or CUDA versions because of floating-point non-determinism in GPU attention kernels and differences in BF16 reduction order. The following specific caveats apply to the released code:

\begin{itemize}[leftmargin=*, topsep=2pt, itemsep=1pt]
  \item The PyTorch DataLoader is configured with \texttt{num\_workers=0} because Windows does not support the multiprocessing fork used by the default worker loop. On Linux, raising the worker count is possible but may shift dataloader-side ordering at sub-batch granularity.
  \item Gradient checkpointing is disabled because its interaction with HyperConnection's ring buffer and the frozen-$K, V$ design produced silent gradient corruption in earlier experiments. Re-enabling it would reduce VRAM at the cost of altered training dynamics.
  \item \texttt{torch.compile} is not used because Triton is unavailable on Windows 11. On Linux, enabling it would yield modest training speedup but has not been validated against the training recipe used here.
\end{itemize}

\section*{Code and Data Availability}

The complete \cart{} implementation, training and evaluation scripts, all figure-generation scripts, and the SQLite database (\texttt{results.db}) of every training run logged in this paper are released at \url{https://github.com/ccapps42/CART}. The database includes per-step train logs (loss, gradient norm, learning rate, spectral radius), per-eval validation perplexities on all three held-out sets, and full configuration provenance for every run, sufficient to regenerate every table and figure in this paper.

\section*{Acknowledgments}

This work was conducted independently, without institutional or external funding.

\section*{Use of Generative AI}

The author used a generative AI language model (Anthropic's Claude) to assist with drafting and editing the manuscript and with writing analysis and figure-generation scripts. All AI-assisted output, including every claim, numerical result, and reference, was reviewed and verified by the author, who takes full responsibility for the content of this paper.

\bibliographystyle{plainnat}
\bibliography{references}

\end{document}